\documentclass[10pt,twocolumn,letterpaper]{article}

\usepackage{cvpr}              % To produce the CAMERA-READY version
\usepackage{graphicx}
\usepackage{amsmath}
\usepackage{amssymb}
\usepackage{booktabs}

\usepackage{epstopdf}
\usepackage{array}
\usepackage{multirow}
\usepackage{booktabs}
\usepackage{balance}
\usepackage[pagebackref,breaklinks,colorlinks]{hyperref}

\usepackage[capitalize]{cleveref}
\crefname{section}{Sec.}{Secs.}
\Crefname{section}{Section}{Sections}
\Crefname{table}{Table}{Tables}
\crefname{table}{Tab.}{Tabs.}

\def\cvprPaperID{1503} % *** Enter the CVPR Paper ID here
\def\confName{CVPR}
\def\confYear{2022}

\makeatletter
\newcommand{\thickhline}{%
    \noalign {\ifnum 0=`}\fi \hrule height 1pt
    \futurelet \reserved@a \@xhline
}

\begin{document}

%%%%%%%%%--------------------- TITLE ---------------------%%%%%%%%%
\title{Deformation and Correspondence Aware Unsupervised Synthetic-to-Real Scene Flow Estimation for Point Clouds}
%%%%%%%%

\author{Zhao Jin$^{1}$ \quad Yinjie Lei$^{1,}$\thanks{Corresponding Author: Yinjie Lei (yinjie@scu.edu.cn)} \quad Naveed Akhtar$^{2}$ \quad Haifeng Li$^{3}$ \quad Munawar Hayat$^{4}$\\
$^{1}$Sichuan University  $^{2}$The University of Western Australia  $^{3}$Central South University  $^{4}$Monash University\\
{\tt\small jinzhao@stu.scu.edu.cn} \quad {\tt\small yinjie@scu.edu.cn} \quad {\tt\small naveed.akhtar@uwa.edu.au}\\
{\tt\small lihaifeng@csu.edu.cn} \quad {\tt\small munawar.hayat@monash.edu}
}

\maketitle

%%%%%%%%%----------------------- ABSTRACT -----------------------%%%%%%%%%
\begin{abstract}
Point cloud scene flow estimation is of practical importance for dynamic scene navigation in autonomous driving. Since scene flow labels are hard to obtain, current methods train their models on synthetic data and transfer them to real scenes. However, large disparities between existing synthetic datasets and real scenes lead to poor model transfer. We make two major contributions to address that. First, we develop a point cloud collector and scene flow annotator for GTA-V engine to automatically obtain diverse realistic training samples without human intervention. With that, we develop a large-scale synthetic scene flow dataset GTA-SF. Second, we propose a mean-teacher-based domain adaptation framework that leverages self-generated pseudo-labels of the target domain. It also explicitly incorporates shape deformation regularization and surface correspondence refinement to address distortions and misalignments in domain transfer. Through extensive experiments, we show that our GTA-SF dataset leads to a consistent boost in model generalization to three real datasets (\textit{i.e.,} Waymo, Lyft and KITTI) as compared to the most widely used FT3D dataset. Moreover, our framework achieves superior adaptation performance on six source-target dataset pairs, remarkably closing the average domain gap by $60 \%$. Data and codes are available at \href{https://github.com/leolyj/DCA-SRSFE}{https://github.com/leolyj/DCA-SRSFE}
% Our code will be released upon the acceptance of this paper.

\end{abstract}

%%%%%%%%%----------------------- Introduction -----------------------%%%%%%%%%
\vspace{-5mm}
\section{Introduction}
\label{sec:intro}
\vspace{-2mm}
Scene flow estimation aims to predict the 3D motion field from two consecutive input frames. 
As a generalization of 2D optical flow, scene flow represents 3D motion of objects and can be used to predict their movement in the future, which is meaningful in robotic navigation and autonomous driving. 
In the early years, scene flow was estimated from stereo or RGB-D images \cite{huguet2007variational, vogel20113d, vogel2013piecewise, wedel2011stereoscopic}.
With the recent advances in 3D sensing and data driven technologies, learning scene flow directly from point clouds has gained significant research attention~\cite{liu2019flownet3d, gu2019hplflownet, puy2020flot, gojcic2021weakly, baur2021slim, jund2021scalable}.

\begin{figure}

\subfloat[Differences between FT3D \cite{mayer2016large} (left) and Waymo \cite{sun2020scalability, jund2021scalable} (right)]{
\centering
\includegraphics[width=0.98\linewidth]{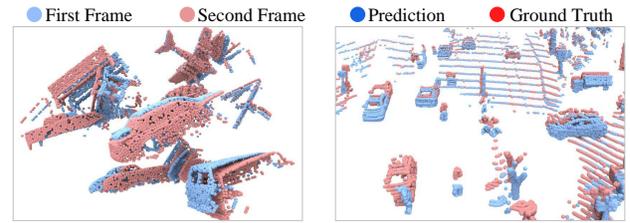}
}
\vspace{0mm}
\quad

\subfloat[Inaccurate predictions (\textit{i.e., } results of adding estimated scene flow to  $1^{\mathrm{st}}$ frame) caused by training on FT3D \cite{mayer2016large} and testing on Waymo\cite{sun2020scalability, jund2021scalable}]{
\centering
\includegraphics[width=0.98\linewidth]{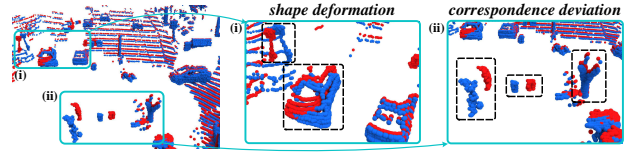}
}
\vspace{-2mm}
\caption{Challenges for Synthetic-to-Real Scene Flow Estimation (SRSFE). (a) Existing synthetic dataset FT3D~\cite{mayer2016large} (left) stacks and moves ShapeNet~\cite{savva2015semantically} objects for data generation, resulting in unnatural scenes distinct from real data (right), \textit{e.g.,}~Waymo \cite{sun2020scalability, jund2021scalable}. (b) Due to domain shift, SRSFE fails to maintain local structure and accurate movement, leading to shape deformation and correspondence deviation. The lack of appropriate synthetic datasets and performance drop in SRSFE motivate our work. 
}
\vspace{-8mm}
\quad
\label{fig:motivation}
\end{figure}

Obtaining training data for scene flow estimation requires 3D motion vector annotation of each point in the scene, which is extremely challenging.
One pragmatic solution is to employ synthetic data for training, whose annotations can be directly generated. 
However, training on synthetic data and testing on real data for scene flow estimation, \textit{i.e.,} Synthetic-to-Real Scene Flow Estimation (SRSFE), faces two major challenges.
First, SRSFE research on point clouds is still in its infancy, and currently there is a lack of synthetic data that adequately captures the real-world dynamics for this task.
The only public synthetic data for SRSFE on point clouds, \textit{i.e.,}~FlyingThings3D \cite{mayer2016large} (FT3D), is generated by randomly moving 3D objects sampled from ShapeNet \cite{savva2015semantically}.  
This simplistic process leads to unnatural scene flows in the data, see Fig.~\ref{fig:motivation}(a).   
Second, SRSFE must overcome the inevitable domain gap caused by the synthetic-to-real setting.
In recent years, extensive studies have been conducted on Unsupervised Domain Adaptation (UDA), which adapts a model to unseen unlabeled data to mitigate the domain gap problem.
However, most of the existing UDA methods \cite{chen2017no, hoffman2016fcns, hoffman2018cycada, ganin2015unsupervised, french2017self, tsai2018learning, tzeng2017adversarial} are designed for 2D tasks to address the domain gap caused by the variance of image texture, color and illumination. 
Much less attention has been paid to UDA  for point clouds \cite{qin2019pointdan, yang2021st3d, yi2021complete, xu2021spg, zhang2021srdan}. This is especially true for SRSFE, for which no systematic study is available to date. 
Compared to other static point cloud tasks, SRSFE has a peculiar requirement of learning correlations between dynamic points. Hence, existing UDA methods are not readily transferable to this task.

In this paper, we address the above two problems.
First, we propose a synthetic point cloud scene flow dataset GTA-V Scene Flow (GTA-SF), to address the lack of dataset. Our data leverages  GTA-V engine~\cite{GTAV} to simulate LiDAR scanning and  autonomously annotate scene flow by aligning the identical entities rendered by the engine.
Compared to FT3D, 
GTA-SF has more realistic scenes and point cloud representation.
Secondly, to bridge the synthetic-to-real domain gap, we propose a UDA framework specifically designed for the SRSFE task.
Observing that `\textit{shape deformation}' and `\textit{correspondence deviation}' are the key contributors to performance degradation in SRSFE - Fig.~\ref{fig:motivation}(b), 
our technique learns deformation and correspondence under a mean-teacher strategy.
We constrain the teacher predictions with rigid shapes and induce a deformation-aware student model to learn desirable scene flow.
To address  correspondence deviation, we leverage object surface relationships to let the model learn better correspondence on  real data.

Our extensive experiments show that our dataset GTA-SF shows remarkable generalization to real-world data, and the proposed framework is highly effective in  reducing the domain gap for the point cloud SRSFE problem. In brief, our contributions can be summarized as follows

\vspace{-2.0mm}
\begin{itemize}
\item We present the first (to the best of our knowledge) systematic study on bridging the domain gap in synthetic to real-world scene flow estimation for point clouds. 
\item We develop a point cloud sequence collector and scene flow annotator for GTA-V engine, and create a large-scale dataset GTA-SF for the SRSFE task. 
\item We propose a mean-teacher domain adaptation framework for point cloud SRSFE that explicitly addresses shape deformations and correspondence deviation. 
\item With extensive experiments, we demonstrate our GTA-SF is closer to real as it enables better performance on real dataset, and our technique consistently surpass common UDA methods across multiple datasets.
\end{itemize}

\begin{figure*}[t]
\noindent
  \centering
  \includegraphics[width=1\linewidth]{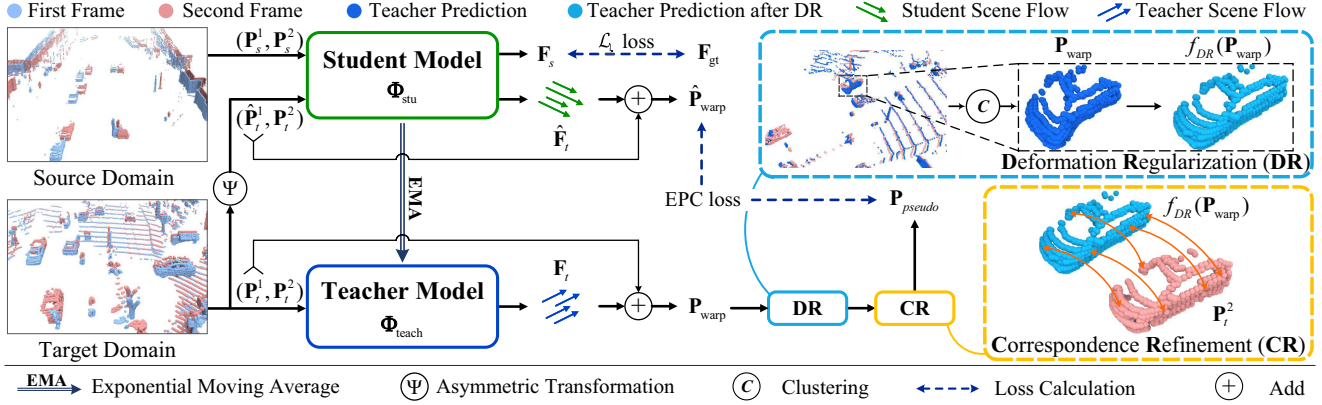}
  \vspace{-6mm}   
  \caption{Framework schematics: Our teacher model $\mathbf{\Phi}_{\text{teach}}$ is updated as the Exponential Moving Average (EMA) of the student model weights $\mathbf{\Phi}_{\text{stu}}$. The first and second frames of the source domain, $\mathbf{P}_{s}^{1}$ and $\mathbf{P}_{s}^{2}$,  are input to the student for which ground truth scene flow $\mathbf{F}_{\mathrm{gt}}$ is available for supervision. The student also expects a transformed version $(\mathbf{\hat{P}}_{t}^{1},\mathbf{P}_{t}^{2})$ of the target domain inputs $(\mathbf{P}_{t}^{1},\mathbf{P}_{t}^{2})$ provided to the teacher. An End-Point Consistency (EPC) is eventually imposed between the scene flow predictions of the teacher $\mathbf{F}_{t}$ and the student $\mathbf{\hat{F}}_{t}$. For EPC, both predictions are first added with the first frame of inputs to obtain $\mathbf{P}_{\mathrm{warp}}$ and $\mathbf{\hat{P}}_{\mathrm{warp}}$. The teacher prediction is subsequently regularized for deformation and refined for correspondence, thereby providing high-quality pseudo-labels $\mathbf{P}_{\textit{pseudo}}$ to compute EPC loss.
  }
  \vspace{-2mm}
  \label{fig:framework}
\end{figure*}

%%%%%%%%%----------------------- Related Work -----------------------%%%%%%%%%

\vspace{-2mm}
\section{Related Works}
\label{sec:rw}
\vspace{-2mm}
\noindent{\bf Scene Flow Estimation on Point Clouds: \ } Scene flow estimation problem  is first introduced and defined as a 3D motion field of points in \cite{vedula1999three}. 
Early works \cite{huguet2007variational, hornacek2014sphereflow, vogel20113d, vogel2013piecewise, vogel20153d, wedel2011stereoscopic} estimate 3D scene flow from  stereo images or RGB-D sequences \cite{huguet2007variational, vogel20113d, vogel2013piecewise, wedel2011stereoscopic}. 
More recently, with the popularity of 3D sensors, an increasing number of techniques  focus on learning scene flow directly from 3D point clouds~\cite{liu2019flownet3d, gu2019hplflownet, puy2020flot, wu2020pointpwc, mittal2020just, wei2021pv, kittenplon2021flowstep3d, gojcic2021weakly, li2021hcrf}. 
For instance, Liu \textit{et al.} \cite{liu2019flownet3d} leveraged  PointNet++~\cite{qi2017pointnet++} for feature extraction and proposed a flow embedding layer for cross-frame geometric relation learning. 
In \cite{gu2019hplflownet}, Gu \textit{et al.}~took  advantage of permutohedral lattice projection and designed a Bilateral Convolutional Layer for two consecutive frames. Optimal transport is utilized to guide scene flow learning in
\cite{puy2020flot, li2021self}. Wei \textit{et al.} \cite{wei2021pv} constructed  point-voxel correlation fields to capture local and long-range relations among points. 
Gojcic \textit{et al.} \cite{gojcic2021weakly} proposed a weakly supervised approach to learn rigid scene flow by only using binary background segmentation and ego-motion annotations. Instances of self-supervised approaches to scene flow estimation can also be found in~\cite{mittal2020just, wu2020pointpwc}. 
Although scene flow estimation from 3D point clouds has shown promises, the domain gap between the real and synthetic data dramatically degrades generalization abilities of the current models in real-world settings. We address this limitation in our proposed framework.

\vspace{0.5mm}
\noindent{\bf Unsupervised Domain Adaptation (UDA): \ } aims to generalize a model trained on a source domain to an unlabeled target domain. The UDA has demonstrated remarkable performance on 2D vision tasks. For 3D point clouds, UDA is also  explored for shape classification \cite{qin2019pointdan, achituve2021self}, semantic segmentation \cite{peng2021sparse, jaritz2020xmuda, yi2021complete, wu2019squeezesegv2} and object detection \cite{wang2020train, yang2021st3d, zhang2021srdan, luo2021unsupervised, xu2021spg}. 
Among these contributions, Qin \textit{et al.} \cite{qin2019pointdan} proposed to learn domain-invariant point cloud representation by global and local feature alignment.
Yi \textit{et al.} \cite{yi2021complete} used  a surface completion network to transform both source and target point clouds into a canonical domain, and trained a shared segmentation network.
Yang \textit{et al.} \cite{yang2021st3d} adopted self-training with memory bank-based pseudo-label generation and curriculum data augmentation for UDA on 3D detection. 
Luo \textit{et al.} \cite{luo2021unsupervised} addressed the problem of inaccurate box-scale by adopting multi-level consistency regularization for the target domain with teacher-student paradigm. 
In general, successful application of UDA  requires addressing task-specific domain shift challenges. For the synthetic-to-real scene flow estimation task, this problem is still unaddressed in the existing literature. Hence, it is the main contribution of this paper.

\vspace{0.5mm}
\noindent{\bf Synthetic-to-Real Transfer Learning: \ } Training with synthetic data is  widely used to avoid laborious annotation process \cite{richter2016playing, richter2017playing, krahenbuhl2018free, johnson2016driving, ros2016synthia, mayer2016large}. Gaming engines, e.g.,~Grand Theft Auto V (GTA-V), have proven useful for generating  synthetic data for various 2D vision tasks, \textit{e.g.}, semantic segmentation \cite{richter2016playing}, optical flow \cite{richter2017playing, krahenbuhl2018free}, object detection \cite{johnson2016driving} and crowd counting \cite{wang2019learning}. Recently, GTA-V has also been used for point cloud data generation for 3D object detection \cite{hurl2019precise}, semantic segmentation \cite{wu2019squeezesegv2} and 3D mesh reconstruction \cite{hu2021sail}. 
For scene flow estimation, the existing synthetic dataset FT3D \cite{mayer2016large} constructs scenes with ShapeNet \cite{savva2015semantically} objects moving along random 3D trajectories. Although useful, significant differences between such scenes and real-world scenarios lead to poor model generalisation on real data. We address it by leveraging the established efficacy of GTA-V to generate more realistic scene flow dataset.

%%%%%%%%%----------------------- Methodology -----------------------%%%%%%%%%

\vspace{-1mm}
\section{Methodology}
\label{sec:method}

\subsection{Problem Formulation}
\label{subsec:prob.def}
\vspace{-2mm}
For our problem, we consider a set of labeled data, $\mathcal{S}=\{(\mathbf{P}_{s}^{i},\mathbf{P}_{s}^{i+1},\mathbf{F}_{\text{gt}}^{i})\mid_{i=1}^{N_{s}}\}$ for the source domain. Here, $\mathbf{P}_{s}^{i}$ and $\mathbf{P}_{s}^{i+1}$ are two successive point cloud frames and $\mathbf{F}_{\text{gt}}^{i}$ is the ground truth scene flow between them. We have $|\mathcal S| = N_{s}$ samples available for the source domain. Correspondingly, we have a set $\mathcal{T}=\{(\mathbf{P}_{t}^{i},\mathbf{P}_{t}^{i+1}, \mathbf{F}_{\text{test}}^{i})\mid_{i=1}^{N_{t}}\}$ for the target domain, for which $\mathbf{P}_{t}^{i}$ and $\mathbf{P}_{t}^{i+1}$ are consecutive frames and $N_{t}$ denotes the number of samples. For $\mathcal{T}$, the scene flow  $\mathbf{F}^i_{\text{test}}$ is unknown. The objective of synthetic-to-real scene flow estimation is to compute an estimator $\Lambda (\mathcal S, \mathbf{P}_{t}^{i}) \rightarrow \mathbf{F}^i_{\text{test}}, \forall~\mathbf{P}_{t}^{i} \in \mathcal T$ such that the source domain is restricted to synthetic data only, and the target domain is the real-world. Due to the large domain shift between the considered $\mathcal S$ and $\mathcal T$, the estimator $\Lambda (.)$ needs to be robust to the domain gap. We adopt the unsupervised domain adaption (UDA) paradigm to address that. The problem is referred to as SRSFE for synthetic-to-real scene flow estimation.

% overview mention 
\vspace{-1.5mm}
\subsection{Technique Overview} 
\vspace{-1.5mm}
\label{subsec:framework}

\noindent
\textbf{Teacher-Student Paradigm }
Schematics of our UDA framework for point cloud SRSFE is provided in Fig.~\ref{fig:framework}.
Our framework employs a student model $\mathbf{\Phi}_{\text{stu}}$ and a teacher model $\mathbf{\Phi}_{\text{teach}}$. 
In the text, we alternatively use these symbols to refer to model weights for brevity.
In our technique, we apply back-propagation to update $\mathbf{\Phi}_{\text{stu}}$, whereas stop gradient is used for  $\mathbf{\Phi}_{\text{teach}}$. For the latter, we use the Exponential Moving Average (EMA) of $\mathbf{\Phi}_{\text{stu}}$ to iteratively update the weights as 
\vspace{-1mm}
\begin{equation}
\mathbf{\Phi}_{\text{teach}}^{\text{updated}}\leftarrow\alpha\mathbf{\Phi}_{\text{teach}}+(1-\alpha)\mathbf{\Phi}_{\text{stu}},
\label{eq:EMA} 
\end{equation}
where $\alpha$ is a smoothing coefficient that dictates the update rate of the teacher model.

\noindent
\textbf{Asymmetric Transformation }
We  encourage domain invariance in the student model by eventually forcing its prediction to match the teacher prediction for an input that is a known transform of the teacher input from the target domain. 
As scene flow is learned from dynamic point cloud sequences, the correlation between two consecutive frames plays a key role in scene flow estimation.
Considering that, we define an asymmetric transformation operation $\Psi$(.,.), for the input  (\textit{i.e.,} two consecutive frames of point clouds).
The operator $\Psi$ stochastically  applies a transformation to the first frame of the input, and leaves the second frame unchanged. 
We consider global rotation and translation for $\Psi(.,.)$, which alter the position without disrupting shapes of objects.
Consider an input with $\mathbf{P}_{t}^{1}$ as the first frame and $\mathbf{P}_{t}^{2}$ as the next frame, the transformation is performed over the input as 
\vspace{-1mm}
\begin{equation}
\Psi(\mathbf{P}_{t}^{1},\mathbf{P}_{t}^{2})=(\mathbf{\hat{P}}_{t}^{1},\mathbf{P}_{t}^{2}),
\end{equation}
where $\mathbf{\hat{P}}_{t}^{1}$ is the transformed version of $\mathbf{P}_{t}^{1}$. 
Through  $\Psi(.,.)$, the models robustly comprehend the notion of correlations between consecutive frames.

\noindent
\textbf{End-Point Consistency }
Ideally, adding scene flow to the first frame should provide an estimate of the second frame. The first frame added by scene flow is referred as `warped frame'. 
We can promote the teacher-student consistency by enforcing the target domain warped frame prediction for the two models to be similar.
For this, we introduce the notion of End-Point Consistency (EPC) between $\mathbf{P}_{\mathrm{warp}}$ and  $\mathbf{\hat{P}}_{\mathrm{warp}}$, such that $\mathbf{P}_{\mathrm{warp}} = \mathbf{P}_{t}^{1}+\mathbf{F}_{t}$ and  $\mathbf{\hat{P}}_{\mathrm{warp}} = \mathbf{\hat{P}}_{t}^{1}+\mathbf{\hat{F}}_{t}$, where $\mathbf{F}_t$ and $\mathbf{\hat{F}}_{t}$ are the predicted scene flows of the teacher and student models respectively.  

As the teacher model provides pseudo labels for the student model, we propose to improve the quality of pseudo labels to teach a better student on the target domain.
To that end, for the teacher prediction, the warped frame is further processed by deformation regularization (\S~\ref{subsec:DR}) to maintain rigid shapes.
We also propose a subsequent correspondence refinement (\S~\ref{subsec:CR}) for better surface alignment of objects.
By forcing the student prediction on the original target domain input (after applying $\Psi$) to be consistent with the teacher model prediction, we effectively encourage domain invariance in the student by enhancing its robustness to input perturbations. It also promotes target domain deformation and correspondence awareness in the student.

\vspace{-1mm}
\subsection{Deformation Regularization (DR)}
\label{subsec:DR}
\vspace{-1mm}
For the SRSFE problem, a model induced over synthetic data must generalize to real-world data. Since synthetic data generally does not faithfully capture real-world details, the model may fail to fully comprehend object shape in the target domain at the desired granularity level. This causes scene flow vectors to have distorted object shapes.

To address the problem, we design a Deformation Regularization (DR) module as a deformation corrector for rigid bodies.
% To address the problem, we design a Deformation Regularization (DR) module to reconstruct rigid shapes of warped point clouds $\mathbf{P}_{\mathrm{warp}}$, thereby maintaining the original object shapes in $\mathbf{P}_{t}^{1}$.
Specifically, for the warped point cloud $\mathbf{P}_{\mathrm{warp}}$ of the teacher model, we first segment it into several distinct clusters $\{\mathbf{C}_{l}\mid_{l=1}^{N_{c}}\}$, where ${N_{c}}$ is the number of clusters.
We then employ the Kabsch algorithm \cite{kabsch1976solution} to estimate a rigid motion $(\mathbf{R}_{l}, \mathbf{t}_{l})$ for each cluster $\mathbf{C}_{l}$ from $\mathbf{P}_{t}^{1}$ to $\mathbf{P}_{\mathrm{warp}}$, where $\mathbf{R}_{l}\in\mathbb{R}^{3\times3}$ and $\mathbf{t}_{l}\in\mathbb{R}^{3}$ denote the rotation and translation matrices. 
The reconstructed cluster $\mathbf{C}_{l}^{\prime}$ are then obtained as
        
\vspace{-1mm}
\begin{equation}
\mathbf{C}_{l}^{\prime}=\left\{ (\mathbf{C}_{l}^{1}\cdot\mathcal{\mathrm{\mathbf{R}}}_{l}+\mathbf{t}_{l})|_{i=1}^{n_{l}}\right\}, 
\end{equation}
where $\mathbf{C}_{l}^{1}$ indicates the corresponding points in $\mathbf{P}_{t}^{1}$ of $\mathbf{C}_{l}$ and $n_{l}$ is the number of points in $\mathbf{C}_{l}$. Let us write that after applying DR, our teacher warped result $\mathbf{P}_{\mathrm{warp}}$ is reconstructed as $f_{DR}(\mathbf{P}_{\mathrm{warp}})=\{\mathbf{C}_{l}^{\prime}|_{l=1}^{N_{c}}\}$. Then, EPC ensures consistency between the student warped results $\mathbf{\hat{P}}_{\mathrm{warp}}$ and $f_{DR}(\mathbf{P}_{\mathrm{warp}})$. This encourages shape distortion awareness in the student model, thereby allowing it to learn adaptive deformations for the target domain. The $f_{DR}(\mathbf{P}_{\mathrm{warp}})$ is later improved with  correspondence refinement.

\begin{figure}[t]
  \centering
  \includegraphics[width=1\linewidth]{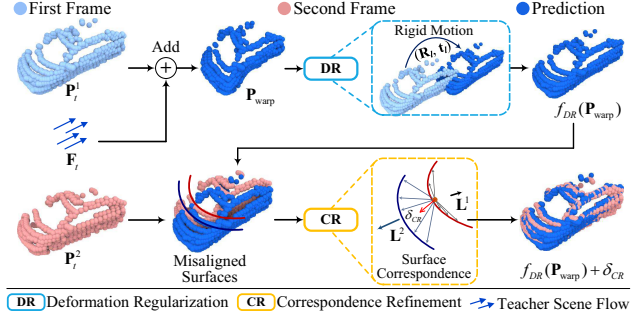}
  \caption{Illustration of the proposed DR and CR to generate high-quality pseudo labels using the teacher model stream for guiding the student. The DR improves object shape and CR improves surface correspondence by refining local geometry. }
  \label{fig:DR_CR}
\vspace{-4mm}
\end{figure}

\vspace{-1mm}
\subsection{Correspondence Refinement (CR)}
\vspace{-1mm}
\label{subsec:CR}
Ideally, a model for SRSFE must be able to maintain correct correspondence between the objects in consecutive frames in the target domain. However, synthetic objects in the source domain have distinct boundaries and geometrically simplified surfaces, whereas the real-world object shapes are much more complex. This is problematic for automatically preserving the desired correspondences when the model is applied in the targeted domain.

To address that, we explicitly encourage cross-frame surface correspondence in our model.
Let us briefly consider the second frame $\mathbf{P}_{t}^{2}$ in an input $(\mathbf{P}_{t}^{1}, \mathbf{P}_{t}^{2}$) as the target frame.
The scene flow $\mathbf{F}_{t}$ identifies  per-point translations that move the first frame $\mathbf{P}_{t}^{1}$ to match  the second frame $\mathbf{P}_{t}^{2}$.
In our setup, if the scene flow vectors are correctly estimated,  the objects in $\mathbf{P}_{\mathrm{warp}}$ and $\mathbf{P}_{t}^{2}$ will have the same  surfaces. 
The accuracy of the estimated scene flow  therefore directly depends on how well the geometric surfaces match across the frames. 
To account for cross-frame surface correspondence, we employ the Laplacian coordinate \cite{sorkine2005laplacian} that records local geometric characteristics of 3D surfaces.
We start with computing the Laplacian coordinate $\mathbf{L}^1$ of each point $p_{\mathrm{warp}}^{j}$ within the warped point cloud $\mathbf{P}_{\mathrm{warp}}$ as
\vspace{-1mm}
\begin{equation}
\mathbf{L}^{1}(p_{\mathrm{warp}}^{j})=\frac{1}{|\mathcal{N}(p_{\mathrm{warp}}^{j},\mathbf{P}_{\mathrm{warp}})|}\sum_{k=1}^{K}\!\!\left(p_{\mathrm{warp}}^{k}\!\!\!-p_{\mathrm{warp}}^{j}\right),
\end{equation}
where $\mathcal{N}(p_{\mathrm{warp}}^{j},\mathbf{P}_{\mathrm{warp}})$ calculates $K$ nearest neighbors of $p_{\mathrm{warp}}^{j}$ in $\mathbf{P}_{\mathrm{warp}}$ and $p_{\mathrm{warp}}^{k}\in\mathcal{N}(p_{\mathrm{warp}}^{j},\mathbf{P}_{\mathrm{warp}})$.
We then extend to cross-frame correspondence by querying neighboring points in the second frame $\mathbf{P}_{t}^{2}$ as
\vspace{-1mm}
\begin{equation}
\mathbf{L}^{2}(p_{\mathrm{warp}}^{j})=\frac{1}{|\mathcal{N}(p_{\mathrm{warp}}^{j},\mathbf{P}_{t}^{2})|}\sum_{k=1}^{K}\left(p_{2}^{k}-p_{\mathrm{warp}}^{j}\right),
\end{equation}
where $\mathcal{N}(p_{\mathrm{warp}}^{j},\mathbf{P}_{t}^{2})$ means neighboring points of $p_{\mathrm{warp}}^{j}$ in $\mathbf{P}_{t}^{2}$ and $p_{2}^{k}\in\mathcal{N}(p_{\mathrm{warp}}^{j},\mathbf{P}_{t}^{2})$. 
The discrepancy between $\mathbf{L}^{1}(p_{\mathrm{warp}}^{j})$ and $\mathbf{L}^{2}(p_{\mathrm{warp}}^{j})$ will provide cues for the surface mis-alignment, which can be utilized to make refinement for pseudo labels of the teacher model.
As individual objects are separated and reconstructed for DR, we further compute refinement vectors for each reconstructed cluster $\mathbf{C}_l^{\prime}$ to maintain the rigid shape as
\vspace{-1mm}
\begin{equation}
\widetilde{\mathbf{C}}_{l}^{\prime}=\mathbf{C}_{l}^{\prime}+\frac{1}{N_{\mathbf{C}_{l}^{\prime}}}\sum_{p_{u}\in \mathbf{C}_{l}^{\prime}}\left(\mathbf{L}^{2}(p_{u})-\mathbf{L}^{1}(p_{u})\right),
\vspace{-2mm}
\end{equation}
where $\widetilde{\mathbf{C}}_{l}$ represents the refined cluster and $N_{\mathbf{C}_{l}^{\prime}}$ is the number of points in cluster $\mathbf{C}_l^{\prime}$. 
After the refinement of individual clusters, we denote the refined warped point clouds as $\mathbf{P}_{pseudo}$, which is the final pseudo labels and  obtained by
\vspace{-2mm}
\begin{equation}
\mathbf{P}_{pseudo}=f_{DR}(\mathbf{P}_{\mathrm{warp}})+\delta_{CR}=\{\widetilde{\mathbf{C}}_{l}^{\prime}|_{l=1}^{N_{c}}\},
\end{equation}
where $\delta_{CR}$ denotes refinement vectors of our CR.
By improving the reconstructed point clouds with CR, teacher scene flow is adjusted for better surface alignment, such that the teacher model stream is able to provide more reliable pseudo labels for the warped point clouds, which are subsequently exploited in student training. We illustrate CR applied after DR for pseudo label improvement in Fig.~\ref{fig:DR_CR}.

\vspace{-0.5mm}
\subsection{Network Training}
\label{subsec:training}
\vspace{-0.5mm}
Our network comprises a student model $\mathbf{\Phi}_{\mathrm{stu}}$ and a teacher model $\mathbf{\Phi}_{\mathrm{teach}}$.
In each iteration, $\mathbf{\Phi}_{\mathrm{stu}}$ is trained with a supervised loss $\mathcal{L}_{\mathrm{source}}$ defined over the source domain and a consistency loss $\mathcal{L}_{\mathrm{EPC}}$ over the target domain.
$\mathcal{L}_{\mathrm{source}}$ is the $\mathcal{L}_{1}$ loss of the between the ground truth scene flow $\mathbf{F}_{\mathrm{gt}}$ and student estimated scene flow $\mathbf{F}_{s}$,~i.e.,
\begin{equation}
\mathcal{L}_{\mathrm{source}}\left(\mathbf{\Phi}_{\mathrm{stu}}\right)=||\mathbf{F}_{s}-\mathbf{F}_{\mathrm{gt}}||_{1}.
\end{equation}

For $\mathcal{L}_{\texttt{EPC}}$, we use the $\mathcal{L}_{1}$ loss between the warped results of the student $\hat{\mathbf{P}}_{\mathrm{warp}}$ and the teacher $\mathbf{P}_{\mathrm{warp}}$, after applying DR and CR, i.e.,
\begin{equation}
\mathcal{L}_{\mathrm{EPC}}\left(\mathbf{\Phi}_{\mathrm{stu}}\right)=||\hat{\mathbf{P}}_{\mathrm{warp}}-(f_{DR}(\mathbf{P}_{\mathrm{warp}})+\delta_{\mathrm{CR}})||_{1}.
\end{equation}
The total loss $\mathcal{L}_{\mathrm{stu}}$ for the student model is given by
\begin{equation}
\mathcal{L}_{\mathrm{stu}}\left(\mathbf{\Phi}_{\mathrm{stu}}\right)=\mathcal{L}_{\mathrm{source}}\left(\mathbf{\Phi}_{\mathrm{stu}}\right)+\mathcal{L}_{\mathrm{EPC}}\left(\mathbf{\Phi}_{\mathrm{stu}}\right).
\end{equation}

For the teacher model, we update its weights $\mathbf{\Phi}_{\mathrm{teach}}$ after each iteration using Eq.~(\ref{eq:EMA}).

\vspace{0.0mm}
\section{GTA-V Scene Flow (GTA-SF) Dataset}
\label{sec:GTA-SF}
\vspace{-1mm}
Another major contribution of this paper is the curation of a large-scale synthetic scene flow dataset, generated using GTA-V~\cite{GTAV}. Below, we first describe our method to collect consecutive LiDAR point clouds and annotate scene flow labels for them automatically in GTA-V engine (\S~\ref{subsec:data_collect}). We then discuss the properties of our dataset in comparison with the existing synthetic scene flow datasets (\S~\ref{subsec:dataset_prop}).

\vspace{-0.5mm}
\subsection{Data Collection}
\label{subsec:data_collect}
\vspace{-0.5mm}
We collect data using GTA-V  engine \cite{GTAV} based on Scrip Hook V \footnote{http://www.dev-c.com/gtav/scripthookv/} and PreSIL \cite{hurl2019precise}. Specifically, we first build a scenario with an autonomous driving car on the road. Then, we attach a synthetic LiDAR collector on the top of the car and collect point clouds at a predefined frequency (\textit{e.g.,} 10Hz). 
With the help of Scrip Hook V, we are able to interact with GTA-V and load properties (\textit{e.g.,} position, belonging entity) of each point.
In order to annotate scene flow vectors, we follow the rigidity assumption of Jund et al.~\cite{jund2021scalable}, and calculate rigid motion for each entity.
% We assign a unique entity ID to every encountered rigid object. 
During the game running, each individual object is assigned a unique entity ID, which is considered a rigid body.
% By loading location $\{x, y, z\}$ and pose $\{\alpha, \beta, \gamma\}$ of each entity $\mathrm{e}_{i}$, we can directly compute scene flow $sf_{i}$ for each point $p_{i}$:
We can directly compute scene flow $\mathbf{f}_{i}$ for each point $\mathbf{p}_{i}$ by loading location $\{x, y, z\}$ and pose $\{\alpha, \beta, \gamma\}$ of its entity $\mathrm{e}_{i}$ as 
\vspace{-2mm}
\begin{equation}
 \mathbf{f}_{i}=\left((\mathbf{p}_{i}-\mathbf{P}_{\mathrm{e}_{i}})\cdot\mathbf{R}_{\mathrm{e}_{i}}^{-1}\cdot\mathbf{\dot{R}}_{\mathrm{e}_{i}}+\mathbf{\dot{P}}_{\mathrm{e}_{i}}\right)-\mathbf{p}_{i},
\end{equation}
where $\mathbf{P}_{\mathrm{e}_{i}}$ and $\mathbf{R}_{\mathrm{e}_{i}}$ are respectively the entity position and rotation matrix in current frame, while $\mathbf{\dot{P}}_{\mathrm{e}_{i}}$ and $\mathbf{\dot{R}}_{\mathrm{e}_{i}}$ are those matrices in the next frame.

For the points without a corresponding entity in the next frame, we compute ego-motion for their scene flow analogous  to entity-motion.
Specifically, the location and pose of the LiDAR are kept consistent with the attached car, and the ego-car motion is computed as the scene flow for the  unmatched entities.
For scene flow estimation, ground points are uninformative. Hence, they are manually removed by existing works with height thresholding~\cite{gu2019hplflownet, liu2019flownet3d}. 
%and are manually removed by simple height thresholding.
Since roads are not always flat, thresholding leads to errors, including undesired removal of foreground object points. 
%ground points are not removed completely and part of foreground objects is lost with ground. 
In  GTA-SF, 
%instead of using height thresholding, 
we systematically remove ground points  by exploiting the entity information, \textit{i.e.,} remove points belonging to the ground entities, which helps in better data quality.

\vspace{-0.5mm}
\subsection{Dataset Properties}
\label{subsec:dataset_prop}
\vspace{-0.5mm}
% \noindent
% \textbf{GTA-V Scene Flow (GTA-SF).}
The proposed GTA-SF is a large-scale synthetic dataset for real-world scene flow estimation. It contains 54,287 pairs of consecutive point clouds with densely annotated scene flow. 
Compared to existing synthetic datasets, GTA-SF collects more realistic point clouds with larger scale, and annotates scene flow beyond point correspondence assumption to fit to physical truth.
In terms of diversity, GTA-SF covers a variety of scenarios including downtown, highway, streets and other driving areas rather than artificial scenes. The point clouds are collected along six different routes pertaining various outdoor areas. 
Moreover, the collected point clouds are high-quality for scene flow learning since the meaningless ground points are carefully removed. We provide more detailed illustrations and quantitative analyses of the properties discussed above in the supplementary material.
% We provide illustrations of the properties discussed above in the supplementary material of the paper. Merits of our dataset are also supported quantitatively in ablation study in \S~\ref{sec:ab_gta}. 

\begin{figure}[!t]
  \centering
%   \fbox{\rule{0pt}{1.8in} \rule{0.9\linewidth}{0pt}}
  \includegraphics[width=1\linewidth]{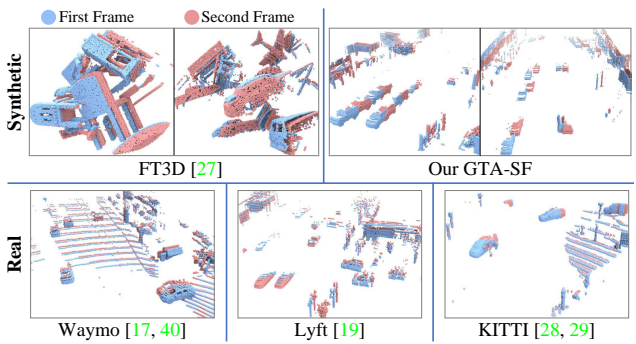}
  \caption{Visual comparisons of synthetic datasets FT3D \cite{mayer2016large} and GTA-SF; and real datasets, Waymo \cite{sun2020scalability, jund2021scalable}, Lyft \cite{Lyft} and KITTI \cite{menze2015object, menze2015joint}. Compared to FT3D, our GTA-SF is more realistic.%NA: @Jin: Also write (our or proposed with GTA-SF in the image).
  }
  \label{fig:ft3d_gta}
  \vspace{-2mm}
\end{figure}

To the best of our knowledge, the FlyingThings3D (FT3D)~\cite{mayer2016large} is the only  widely used synthetic dataset for point cloud scene flow estimation.
It contains 19,640 training and 3,824 testing samples, which makes it smaller in size as compared to our GTA-SF. It builds scenes by stacking 3D objects from ShapeNet \cite{savva2015semantically} and randomly moving them between two frames. Whereas effective, this strategy is unnatural for real-world scenes. 
Tab.~\ref{tab:gta_ft3d} shows a brief comparison between GTA-SF and FT3D. 
Fig.~\ref{fig:ft3d_gta} shows a visual comparison between FT3D and our GTA-SF. We also provide more comprehensive illustrations in the supplementary material.
Our empirical evaluation in \S~\ref{subsec:ex_results} also verifies that GTA-SF considerably narrows down the domain gap between the synthetic and real-world data.

\begin{table}[!t]
\caption{Comparison between FT3D~\cite{mayer2016large} and our GTA-SF.}
\vspace{-2mm}
\centering{}%
\begin{tabular}{>{\centering}m{19mm}>{\centering}m{11mm}>{\centering}m{12mm}>{\centering}m{23mm}}
% \hline 
\thickhline
\multirow{1}{19mm}{\centering{}{\footnotesize{}}} & {\footnotesize{}Frames} & {\footnotesize{}Label} & {\footnotesize{}Scenes}\tabularnewline
\hline 
\multirow{1}{19mm}{\centering{}{\footnotesize{}FT3D~\cite{mayer2016large}}} & {\footnotesize{}23,464} & {\footnotesize{}Unreal} & {\footnotesize{}Objects Stacking}\tabularnewline
\multirow{1}{19mm}{\centering{}{\footnotesize{}GTA-SF (Ours)}}  & {\footnotesize{}54,287} & {\footnotesize{}Realistic} & {\footnotesize{}Vehicle Driving}\tabularnewline
% \hline 
\thickhline
\end{tabular}
\label{tab:gta_ft3d}
\vspace{-2mm}
\end{table}

\vspace{0.0mm}
\section{Experiments}
\label{sec:experiments}
\vspace{-1mm}

\subsection{Real-World Datasets}
\label{subsec:datasets}
\vspace{-1mm}
\noindent
\textbf{Waymo.} The Waymo Open Dataset (WOD) \cite{sun2020scalability} contains 158,081 training and 39,987 validation frames of point clouds with 3D object detection annotations, captured in real-world.
 \cite{jund2021scalable} expand WOD with scene flow annotations by using the tracked bounding boxes of objects.
% In our experiments, we use the training set to train the UDA framework and report the testing results on the validation set.
Note that Waymo compensates for ego-motion before computing scene flow.
It only considers the absolute movements of objects and the annotations for stationary object points are set to zero.
However, previous datasets \cite{mayer2016large, menze2015object, menze2015joint} make no compensation for ego-motion.
The absolute movement and the relative movement caused by ego-motion are estimated simultaneously.
For fair comparisons, we follow previous datasets and retrieve ego-motion vectors for the points based on pose information provided in~WOD.
% remove ground, distance>60

\vspace{0.5mm}
\noindent
\textbf{Lyft.} The Lyft Level 5 dataset \cite{Lyft} consists of 18,900 training and 3,780 validation frames of LiDAR point clouds. It is usually used for 3D object detection and does not provide scene flow labels. As the detection labels and sensor parameters are given, we follow \cite{jund2021scalable} and generate scene flow labels for Lyft without compensating for ego-motion. %In consistent with Waymo, the training frames participate in UDA training and the validation frames are used for evaluation.

\vspace{0.5mm}
\noindent
\textbf{KITTI.} KITTI Scene Flow 2015 \cite{menze2015object, menze2015joint} is a popular benchmark for scene flow estimation. We follow the pre-processing steps in \cite{gu2019hplflownet} to generate point clouds with scene flow annotation, which contains 142 pairs of frames.
% and remove ground points with height ($<0.3m$).

\vspace{-1mm}
\subsection{Evaluation Setup}
\label{subsec:ex_setups}
\vspace{-1mm}

\noindent
\textbf{Implementation Details.} For real-world datasets, we first transform them into the same coordinate system and remove ground points with height $<0.3m$. For our framework, see the supplementary material for implementation details.

%%%%%%%%%%%%%%%%%%%%%%%%%%%%%%%%%%%% main experiments %%%%%%%%%%%%%%%%%%%%%%%%

\renewcommand\arraystretch{0.8}
%%%%%%%%%%%%%%%%%%%%%%%%%%%%%%%%%%%%%%%%%%%%%% datasets %%%%%%%%%%%%%%%%%%%%
\begin{table}[!t]
\caption{Testing EPE3D ($m$) of existing methods pretrained on different source datasets on Waymo \cite{sun2020scalability, jund2021scalable}. }
\vspace{-2mm}

\centering{}%
\begin{tabular}{>{\centering}m{20mm}|>{\centering}m{11.6mm}>{\centering}m{11.9mm}>{\centering}m{8.2mm}>{\centering}m{10.5mm}}
% \hline 
\thickhline
\multirow{2}{20mm}{\centering{}{\scriptsize{}Source$\rightarrow$Target}}  & {\scriptsize{}FlowNet3D} & {\scriptsize{}HPLFlowNet} & {\scriptsize{}FLOT} & {\scriptsize{}PV-RAFT}\tabularnewline
\centering{} & {\scriptsize{}\cite{liu2019flownet3d}} & {\scriptsize{}\cite{gu2019hplflownet}} & {\scriptsize{}\cite{puy2020flot}} & {\scriptsize{}\cite{wei2021pv}}\tabularnewline
\hline 
{\scriptsize{}FT3D$\rightarrow$Waymo} & {\scriptsize{}0.3546} & {\scriptsize{}0.2505} & {\scriptsize{}0.3299} & {\scriptsize{}0.2621}\tabularnewline
{\scriptsize{}GTA-SF$\rightarrow$Waymo} & {\scriptsize{}0.1589} & {\scriptsize{}0.1146} & {\scriptsize{}0.1081} & {\scriptsize{}0.0585}\tabularnewline
{\scriptsize{}Waymo$\rightarrow$Waymo} & {\scriptsize{}0.1067} & {\scriptsize{}0.0501} & {\scriptsize{}0.0680} & {\scriptsize{}0.0433}\tabularnewline
% \hline 
\thickhline

\end{tabular}
\vspace{-1.0em}
\label{tab:datasets}
\end{table}

% \linespread{1}
\renewcommand\arraystretch{1.0}
%%%%%%%%%%%%%%%%%%%%%%%%%%%%%%%%%%%%%%%%%%%%%% datasets %%%%%%%%%%%%%%%%%%%%

\renewcommand\arraystretch{0.9}
\begin{table*}[!t]
% \linespread{0.9}
\caption{Performance comparison on six source-target pairs. HPLFlowNet \cite{gu2019hplflownet} is used as the baseline directly transferred from source to target. Comparison considers Synthetic-to-Real (S$\rightarrow$R) and Real-to-Real (R$\rightarrow$R) transfer. EPE ($m$), AS ($\%$), AR ($\%$) and Out ($\%$) are scene flow estimation evaluation metrics. 
`-' indicates Oracle results on KITTI are not available since no training data is provided. 
Best results for S$\rightarrow$R are in bold. $\downarrow$ and $\uparrow$ respectively indicate negative and positive polarity. E, D and C mean EPC, DR and CR respectively. 
}
% \linespread{1.0}
\vspace{-1mm}
\centering{}%
\begin{tabular}{>{\centering}m{5mm}>{\centering}m{21mm}>{\centering}m{7mm}>{\centering}m{6.5mm}>{\centering}m{6.5mm}>{\centering}m{6.5mm}>{\centering}m{0mm}>{\centering}m{7mm}>{\centering}m{6.5mm}>{\centering}m{6.5mm}>{\centering}m{6.5mm}>{\centering}m{0mm}>{\centering}m{7mm}>{\centering}m{6.5mm}>{\centering}m{6.5mm}>{\centering}m{6.5mm}}
% \hline
\thickhline
% \toprule[0.1em] 
 & \multirow{2}{21mm}{{\footnotesize{}Methods}}  & \multicolumn{4}{c}{\textbf{\footnotesize{}GTA-SF$\longrightarrow$Waymo}} & & \multicolumn{4}{c}{\textbf{\footnotesize{}GTA-SF$\longrightarrow$Lyft}} & & \multicolumn{4}{c}{\textbf{\footnotesize{}GTA-SF$\longrightarrow$KITTI}}\tabularnewline
% \cmidrule(lr){3-6} \cmidrule(lr){7-10} \cmidrule(lr){11-14}
\cline{3-6} \cline{8-11} \cline{13-16}   
 & & {\footnotesize{}EPE$\downarrow$} & {\footnotesize{}AS$\uparrow$} & {\footnotesize{}AR$\uparrow$} & {\footnotesize{}Out$\downarrow$} & &  {\footnotesize{}EPE$\downarrow$} & {\footnotesize{}AS$\uparrow$} & {\footnotesize{}AR$\uparrow$} & {\footnotesize{}Out$\downarrow$} & &  {\footnotesize{}EPE$\downarrow$} & {\footnotesize{}AS$\uparrow$} & {\footnotesize{}AR$\uparrow$} & {\footnotesize{}Out$\downarrow$}\tabularnewline
\hline 
% \hline
% \midrule[0.05em] 
\multirow{6}{5mm}{\centering{}{\footnotesize{}S$\rightarrow$R}} & \centering{}{\footnotesize{}Baseline \cite{gu2019hplflownet}} & {\footnotesize{}0.1061} & {\footnotesize{}32.35} & {\footnotesize{}66.21} & {\footnotesize{}65.35} & & {\footnotesize{}0.1802} & {\footnotesize{}28.36} & {\footnotesize{}73.66} & {\footnotesize{}34.93} & & {\footnotesize{}0.0932} & {\footnotesize{}52.29} & {\footnotesize{}81.39} & {\footnotesize{}33.75}\tabularnewline
 & \centering{}{\footnotesize{}MMD \cite{long2013transfer}} & {\footnotesize{}0.1068} & {\footnotesize{}37.24} & {\footnotesize{}68.85} & {\footnotesize{}66.91} & & {\footnotesize{}0.1563} & {\footnotesize{}41.91} & {\footnotesize{}78.51} & {\footnotesize{}31.35} & & {\footnotesize{}0.0877} & {\footnotesize{}45.87} & {\footnotesize{}79.45} & {\footnotesize{}41.69}\tabularnewline
 & \centering{}{\footnotesize{}Self-Ensemble \cite{french2017self}} & {\footnotesize{}0.0981} & {\footnotesize{}44.05} & {\footnotesize{}71.05} & {\footnotesize{}62.10} & & {\footnotesize{}0.1681} & {\footnotesize{}32.80} & {\footnotesize{}75.59} & {\footnotesize{}34.05} & & {\footnotesize{}0.0869} & {\footnotesize{}51.11} & {\footnotesize{}79.18} & {\footnotesize{}37.49}\tabularnewline
 & \centering{}{\footnotesize{}Ours(E)} & {\footnotesize{}0.0894} & {\footnotesize{}38.68} & {\footnotesize{}75.68} & {\footnotesize{}61.68} & & {\footnotesize{}0.1506} & {\footnotesize{}38.75} & {\footnotesize{}80.22} & {\footnotesize{}29.06} & & {\footnotesize{}0.0848} & {\footnotesize{}51.01} & {\footnotesize{}81.18} & {\footnotesize{}37.52}\tabularnewline
 & \centering{}{\footnotesize{}Ours(E+D)} & {\footnotesize{}0.0887} & {\footnotesize{}40.60} & {\footnotesize{}77.30} & {\footnotesize{}60.70} & & {\footnotesize{}0.1454} & {\footnotesize{}41.11} & {\footnotesize{}81.49} & {\footnotesize{}29.06} & & {\footnotesize{}0.0748} & {\footnotesize{}54.11} & {\footnotesize{}86.53} & {\footnotesize{}31.58}\tabularnewline
 & \centering{}{\footnotesize{}Ours(E+D+C)} & \textbf{\footnotesize{}0.0683} & \textbf{\footnotesize{}58.57} & \textbf{\footnotesize{}87.98} & \textbf{\footnotesize{}47.40} & & \textbf{\footnotesize{}0.1277} & \textbf{\footnotesize{}56.35} & \textbf{\footnotesize{}85.50} & \textbf{\footnotesize{}24.62} & & \textbf{\footnotesize{}0.0464} & \textbf{\footnotesize{}80.53} & \textbf{\footnotesize{}96.85} & \textbf{\footnotesize{}18.75}\tabularnewline
\hline 
% \hline
% \bottomrule[0.03em]
% \toprule[0.03em]
{\centering{}{\footnotesize{}R$\rightarrow$R}} & \centering{}{\footnotesize{}Oracle} & {\footnotesize{}0.0501} & {\footnotesize{}74.82} & {\footnotesize{}92.20} & {\footnotesize{}40.88} & & {\footnotesize{}0.1058} & {\footnotesize{}68.92} & {\footnotesize{}87.04} & {\footnotesize{}22.68} & & {\footnotesize{}-} & {\footnotesize{}-} & {\footnotesize{}-} & {\footnotesize{}-}\tabularnewline
%  & \multirow{1}{13mm}{\centering{}{\footnotesize{}Closed Gap}} & {\footnotesize{}67.50\%} & {\footnotesize{}61.74\%} & {\footnotesize{}83.76\%} & {\footnotesize{}73.36\%} & {\footnotesize{}70.56\%} & {\footnotesize{}69.01\%} & {\footnotesize{}88.49\%} & {\footnotesize{}84.16\%} & {\footnotesize{}50.21\%} & {\footnotesize{}54.01\%} & {\footnotesize{}18.99\%} & {\footnotesize{}44.44\%}\tabularnewline
\hline 
\hline
% \midrule[0.1em]
 & \multirow{2}{21mm}{{\footnotesize{}Methods}}  &
 \multicolumn{4}{c}{\textbf{\footnotesize{}FT3D$\longrightarrow$Waymo}} & & \multicolumn{4}{c}{\textbf{\footnotesize{}FT3D$\longrightarrow$Lyft}} & & \multicolumn{4}{c}{\textbf{\footnotesize{}FT3D$\longrightarrow$KITTI}}\tabularnewline
\cline{3-6} \cline{8-11} \cline{13-16}  
% \cmidrule(lr){3-6} \cmidrule(lr){7-10} \cmidrule(lr){11-14}
& & {\footnotesize{}EPE$\downarrow$} & {\footnotesize{}AS$\uparrow$} & {\footnotesize{}AR$\uparrow$} & {\footnotesize{}Out$\downarrow$} & & {\footnotesize{}EPE$\downarrow$} & {\footnotesize{}AS$\uparrow$} & {\footnotesize{}AR$\uparrow$} & {\footnotesize{}Out$\downarrow$} & & {\footnotesize{}EPE$\downarrow$} & {\footnotesize{}AS$\uparrow$} & {\footnotesize{}AR$\uparrow$} & {\footnotesize{}Out$\downarrow$}\tabularnewline
\hline 
% \midrule[0.05em] 
\multirow{6}{5mm}{\centering{}{\footnotesize{}S$\rightarrow$R}} & \centering{}{\footnotesize{}Baseline \cite{gu2019hplflownet}} & {\footnotesize{}0.2477} & {\footnotesize{}31.59} & {\footnotesize{}57.22} & {\footnotesize{}77.08} & & {\footnotesize{}0.8486} & {\footnotesize{}13.18} & {\footnotesize{}30.42} & {\footnotesize{}79.10} & & {\footnotesize{}0.1169} & {\footnotesize{}47.83} & {\footnotesize{}77.76} & {\footnotesize{}41.03}\tabularnewline
 & \centering{}{\footnotesize{}MMD \cite{long2013transfer}} & {\footnotesize{}0.2179} & {\footnotesize{}24.12} & {\footnotesize{}55.09} & {\footnotesize{}79.52} & & {\footnotesize{}0.7158} & {\footnotesize{}10.48} & {\footnotesize{}29.21} & {\footnotesize{}80.05} & & {\footnotesize{}0.1165} & {\footnotesize{}37.42} & {\footnotesize{}78.46} & {\footnotesize{}42.75}\tabularnewline
 & \centering{}{\footnotesize{}Self-Ensemble \cite{french2017self}} & {\footnotesize{}0.2342} & {\footnotesize{}33.20} & {\footnotesize{}55.54} & {\footnotesize{}78.72} & & {\footnotesize{}0.7366} & {\footnotesize{}13.20} & {\footnotesize{}32.52} & {\footnotesize{}77.23} & & {\footnotesize{}0.1166} & {\footnotesize{}41.88} & {\footnotesize{}77.15} & {\footnotesize{}44.11}\tabularnewline
 & \centering{}{\footnotesize{}Ours(E)} & {\footnotesize{}0.2339} & {\footnotesize{}28.44} & {\footnotesize{}55.76} & {\footnotesize{}77.84} & & {\footnotesize{}0.7330} & {\footnotesize{}10.22} & {\footnotesize{}29.24} & {\footnotesize{}80.35} & & {\footnotesize{}0.1193} & {\footnotesize{}40.50} & {\footnotesize{}75.75} & {\footnotesize{}46.25}\tabularnewline
 & \centering{}{\footnotesize{}Ours(E+D)} & {\footnotesize{}0.2091} & {\footnotesize{}29.56} & {\footnotesize{}56.18} & {\footnotesize{}78.97} & & {\footnotesize{}0.5092} & {\footnotesize{}13.25} & {\footnotesize{}35.61} & {\footnotesize{}74.52} & & {\footnotesize{}0.0992} & {\footnotesize{}46.86} & {\footnotesize{}81.95} & {\footnotesize{}37.83}\tabularnewline
 & \centering{}{\footnotesize{}Ours(E+D+C)} & \textbf{\footnotesize{}0.1251} & \textbf{\footnotesize{}48.87} & \textbf{\footnotesize{}78.40} & \textbf{\footnotesize{}57.29} & & \textbf{\footnotesize{}0.4442} & \textbf{\footnotesize{}25.90} & \textbf{\footnotesize{}51.61} & \textbf{\footnotesize{}58.59} & & \textbf{\footnotesize{}0.0516} & \textbf{\footnotesize{}79.37} & \textbf{\footnotesize{}96.81} & \textbf{\footnotesize{}18.04}\tabularnewline
\hline 
% \hline 
{\centering{}{\footnotesize{}R$\rightarrow$R}} & \centering{}{\footnotesize{}Oracle} & {\footnotesize{}0.0501} & {\footnotesize{}74.82} & {\footnotesize{}92.20} & {\footnotesize{}40.88} & & {\footnotesize{}0.1058} & {\footnotesize{}68.92} & {\footnotesize{}87.04} & {\footnotesize{}22.68} & & {\footnotesize{}-} & {\footnotesize{}-} & {\footnotesize{}-} & {\footnotesize{}-}
\tabularnewline
%  & \multirow{1}{13mm}{\centering{}{\footnotesize{}Closed Gap}} & {\footnotesize{}62.04\%} & {\footnotesize{}39.97\%} & {\footnotesize{}60.55\%} & {\footnotesize{}54.67\%} & {\footnotesize{}54.44\%} & {\footnotesize{}22.82\%} & {\footnotesize{}37.42\%} & {\footnotesize{}36.35\%} & {\footnotesize{}55.86\%} & {\footnotesize{}65.94\%} & {\footnotesize{}24.50\%} & {\footnotesize{}56.03\%}\tabularnewline
% \hline 
\thickhline
% \bottomrule[0.1em]
\label{tab:DA}
\end{tabular}
\vspace{-1.7em}
\end{table*}

% \linespread{1.0}
%%%%%%%%%%%%%%%%%%%%%%%%%%%%%%%%%%%% main experiments %%%%%%%%%%%%%%%%%%%%%%%%

\vspace{0.5mm}
\noindent
\textbf{Comparison Methods.} 
%To validate the merit of our proposed domain adaptation framework,
We provide comparison with the following methods. (1) \textit{Baseline} indicates pretraining the model on source domain and directly evaluating it on the target domain. (2) \textit{MMD} \cite{long2013transfer} adopts Maximum Mean Discrepancy (MMD) for cross-domain feature alignment. (3) \textit{Self-Ensemble} \cite{french2017self} adopts mean-teacher with $\mathcal{L}_{1}$ loss between student and teacher estimated scene flow vectors, and (4)~\textit{Oracle} trains fully-supervised model on  target domain.

\vspace{0.5mm}
\noindent
\textbf{Evaluation Metrics.} Following  \cite{gu2019hplflownet, liu2019flownet3d, wei2021pv, puy2020flot}, we adopt four evaluation metrics. The metrics are calculated between estimated scene flow $\mathbf{F}$ and ground truth $\mathbf{F}_{\mathrm{gt}}$.
% Let us denote the estimated scene flow and  ground truth as $\mathbf{F}$ and $\mathbf{F}_{\mathrm{gt}}$,  respectively. The metrics are defined as follows.

\vspace{0.5mm}
\noindent
\textit{EPE3D (\textbf{EPE})} ($m$): $||\mathbf{F}-\mathbf{F}_{\mathrm{gt}}||_2$ computes the $l_2$ distance between the estimated and ground truth scene flow vectors.

\noindent
\textit{ACC Strict (\textbf{AS})} ($\%$): is the percentage of points with EPE3D $< 0.05m$ or relative error $< 5\%$.

\noindent
\textit{ACC Relax (\textbf{AR})} ($\%$): is the percentage of points with EPE3D $< 0.1m$ or relative error $< 10\%$.

\noindent
\textit{Outliers (\textbf{Out})} ($\%$): is the percentage of points with EPE3D $> 0.3m$ or relative error $> 10\%$.

\noindent

%%%%%%%%%%%%%%%%%%%%%%%%%%%%%%%%%%%%%%%%%%%%%%%%%%%% PV-RAFT %%%%%%%%%%%%%%%%%%%
% \linespread{0.85}
% \renewcommand\arraystretch{0.96}
\renewcommand\arraystretch{1}
\begin{table}[!t]
\caption{Comparisons of PV-RAFT \cite{wei2021pv} baseline from synthetic datasets to Waymo. Our UDA framework achieves remarkable performance in closing the domain gap for PV-RAFT. $\downarrow$ and $\uparrow$ respectively indicate negative and positive polarity.}
% \linespread{1.0}
\vspace{-1mm}
\centering{}%
\begin{tabular}{>{\raggedright}m{8mm}>{\raggedright}m{15mm}>{\centering}m{8mm}>{\centering}m{8mm}>{\centering}m{8mm}>{\centering}m{8mm}}
% \hline 
\thickhline
\multicolumn{2}{c}{{\footnotesize{}Methods}} & {\footnotesize{}EPE$\downarrow$} & {\footnotesize{}AS$\uparrow$} & {\footnotesize{}AR$\uparrow$} & {\footnotesize{}Out$\downarrow$}\tabularnewline
\multicolumn{6}{c}{\textbf{\footnotesize{}GTA-SF$\rightarrow$Waymo}}\tabularnewline
\hline 
% \cmidrule(r){1-2} \cmidrule(l){3-6}
% \hline 
\multirow{2}{8mm}{\centering{}{\footnotesize{}S$\rightarrow$R}} & \multirow{1}{15mm}{\centering{}{\footnotesize{}Baseline \cite{wei2021pv}}} & {\footnotesize{}0.0585} & {\footnotesize{}71.38} & {\footnotesize{}90.74} & {\footnotesize{}42.15}\tabularnewline
 & \multirow{1}{15mm}{\centering{}{\footnotesize{}Ours}} & \textbf{\footnotesize{}0.0474} & \textbf{\footnotesize{}79.93} & \textbf{\footnotesize{}94.14} & \textbf{\footnotesize{}35.61}\tabularnewline
\hline 
% \hline 
% \cline{1-2} 
% \cmidrule(r){1-2}
\centering{}{\footnotesize{}R$\rightarrow$R} & \centering{}{\footnotesize{}Oracle} & {\footnotesize{}0.0433} & {\footnotesize{}84.70} & {\footnotesize{}95.07} & {\footnotesize{}33.09}\tabularnewline
%  & \centering{}{\footnotesize{}Closed Gap} & {\footnotesize{}72.50\%} & {\footnotesize{}64.18\%} & {\footnotesize{}78.57\%} & {\footnotesize{}72.20\%}\tabularnewline
\hline 
\hline
\multicolumn{6}{c}{\textbf{\footnotesize{}FT3D$\rightarrow$Waymo}}\tabularnewline
\hline 
\multirow{2}{8mm}{\centering{}{\footnotesize{}S$\rightarrow$R}} & \multirow{1}{15mm}{\centering{}{\footnotesize{}Baseline \cite{wei2021pv}}} & {\footnotesize{}0.2620} & {\footnotesize{}43.59} & {\footnotesize{}68.25} & {\footnotesize{}63.21}\tabularnewline
 & \multirow{1}{15mm}{\centering{}{\footnotesize{}Ours}} & \textbf{\footnotesize{}0.1219} & \textbf{\footnotesize{}62.29} & \textbf{\footnotesize{}82.53} & \textbf{\footnotesize{}47.71}\tabularnewline
% \hline 
\hline 
% \cline{1-2} 
% \cmidrule(r){1-2}
\centering{}{\footnotesize{}R$\rightarrow$R} & \centering{}{\footnotesize{}Oracle} & {\footnotesize{}0.0433} & {\footnotesize{}84.70} & {\footnotesize{}95.07} & {\footnotesize{}33.09}\tabularnewline
%  & \centering{}{\footnotesize{}Closed Gap} & {\footnotesize{}62.04\%} & {\footnotesize{}39.97\%} & {\footnotesize{}60.55\%} & {\footnotesize{}54.67\%}\tabularnewline
% \hline
\thickhline

\end{tabular}
% \vspace{-0.8em}
\vspace{-1.2em}
\label{tab:pv-raft}
\end{table}
%%%%%%%%%%%%%%%%%%%%%%%%%%%%%%%%%%%%%%%%%%%%%%%%%%%% PV-RAFT %%%%%%%%%%%%%%%%%%%

\vspace{-1mm}
\subsection{Experimental Results}
\label{subsec:ex_results}
\vspace{-1mm}
\noindent
\textbf{Comparison of Synthetic Datasets.}
We first verify the existence of synthetic-to-real domain gap by evaluating recent methods FlowNet3D \cite{liu2019flownet3d}, HPLFlowNet \cite{gu2019hplflownet}, FLOT \cite{puy2020flot} and PV-RAFT \cite{wei2021pv}. We train them on three datasets: FT3D \cite{mayer2016large}, GTA-SF and Waymo \cite{sun2020scalability, jund2021scalable}, and then evaluate them on Waymo. Our experimental results in Tab.~\ref{tab:datasets} show that the models trained on FT3D face a serious performance gap, as  compared to direct training on Waymo. It verifies the existence of large domain gap between synthetic and real datasets. Comparing FT3D and GTA-SF, we see that GTA-SF$\rightarrow$Waymo has a much smaller performance gap. Similar trends are apparent on Lyft and KITTI in Tab.~\ref{tab:DA},  establishing that our GTA-SF is more compatible to real data.

\vspace{0.5mm}
\noindent
\textbf{Synthetic-to-Real Transfer.} 
In Tab.~\ref{tab:DA}, we compare our domain adaptation method for synthetic-to-real scene flow estimation with HPLFlowNet \cite{gu2019hplflownet} Baseline, Oracle and two general-purpose UDA methods (\textit{i.e.,} MMD \cite{long2013transfer} and Self-Ensemble \cite{french2017self}). The results on six source-target pairs demonstrate the superior performance of our method, and its capability to largely close the performance gap between Baseline and Oracle by $60\%$ in EPE. Note that the Oracle results on KITTI are not available since no training data is provided, we also achieve 55.86\% improvement in EPE. Compared with general-purpose UDA methods, our framework  surpasses them on all four evaluation metrics because our technique enables scene flow estimator to be deformation and correspondence aware in the target domain. Also, compared to FT3D, we observe a smaller performance gap for the GTA-SF transferred models.

Our framework also shows remarkable compatibility with other mainstream scene flow estimators. 
As shown in Tab.~\ref{tab:pv-raft} and Tab.~\ref{tab:flot}, it achieves superior performance in closing the domain gap for PV-RAFT \cite{wei2021pv} and FLOT \cite{puy2020flot}, which is consistent with HPLFlowNet \cite{gu2019hplflownet}.
Tab.~\ref{tab:pv-raft} shows the domain adaptation performance of our framework with PV-RAFT on GTA-SF$\rightarrow$Waymo and  FT3D$\rightarrow$Waymo. 
We can narrow the performance gap by 62.04\% to 72.50\% on EPE. % on the used metrics.
% The EPE is only 0.0474$m$ on GTA-SF$\rightarrow$Waymo, which is close to the Oracle result (0.0433$m$). 
% The EPE on GTA-SF$\rightarrow$Waymo (0.0474) is even close to the Oracle result (0.0433).
In Tab.~\ref{tab:flot}, the results on FLOT also show similar trends.

%%%%%%%%%%%%%%%%%%%%%%%%%%%%%%%%%%%%%%%%%%%%%%%%%%%% FLOT %%%%%%%%%%%%%%%%%%%
\begin{table}[t]
% \linespread{0.85}
\caption{Comparisons of FLOT \cite{puy2020flot} baseline from synthetic datasets to Waymo. Our UDA framework is compatible with FLOT and shows consistent performance in closing domain gap. $\downarrow$ and $\uparrow$ respectively indicate negative and positive polarity.}
% \linespread{1.0}
\vspace{-1mm}
\centering{}%
\begin{tabular}{>{\raggedright}m{8mm}>{\raggedright}m{15mm}>{\centering}m{8mm}>{\centering}m{8mm}>{\centering}m{8mm}>{\centering}m{8mm}}
% \hline 
\thickhline
\multicolumn{2}{c}{{\footnotesize{}Methods}} & {\footnotesize{}EPE$\downarrow$} & {\footnotesize{}AS$\uparrow$} & {\footnotesize{}AR$\uparrow$} & {\footnotesize{}Out$\downarrow$}\tabularnewline
\multicolumn{6}{c}{\textbf{\footnotesize{}GTA-SF$\rightarrow$Waymo}}\tabularnewline
% \hline 
\hline 
% \cmidrule(r){1-2} \cmidrule(l){3-6}
% \hline 
\multirow{2}{8mm}{\centering{}{\footnotesize{}S$\rightarrow$R}} & \multirow{1}{15mm}{\centering{}{\footnotesize{}Baseline \cite{puy2020flot}}} & {\footnotesize{}0.1081} & {\footnotesize{}45.36} & {\footnotesize{}75.72} & {\footnotesize{}57.41}\tabularnewline
 & \multirow{1}{15mm}{\centering{}{\footnotesize{}Ours}} & \textbf{\footnotesize{}0.0888} & \textbf{\footnotesize{}59.15} & \textbf{\footnotesize{}82.58} & \textbf{\footnotesize{}49.96}\tabularnewline
% \hline 
\hline
% \cline{1-2} 
% \cmidrule(r){1-2}
\centering{}{\footnotesize{}R$\rightarrow$R} & \centering{}{\footnotesize{}Oracle} & {\footnotesize{}0.0680} & {\footnotesize{}72.78} & {\footnotesize{}89.66} & {\footnotesize{}41.94}\tabularnewline
\hline 
\hline 
\multicolumn{6}{c}{\textbf{\footnotesize{}FT3D$\rightarrow$Waymo}}\tabularnewline
\hline 
\multirow{2}{8mm}{\centering{}{\footnotesize{}S$\rightarrow$R}} & \multirow{1}{15mm}{\centering{}{\footnotesize{}Baseline \cite{puy2020flot}}} & {\footnotesize{}0.3299} & {\footnotesize{}27.07} & {\footnotesize{}48.20} & {\footnotesize{}78.63}\tabularnewline
 & \multirow{1}{15mm}{\centering{}{\footnotesize{}Ours}} & \textbf{\footnotesize{}0.1432} & \textbf{\footnotesize{}52.19} & \textbf{\footnotesize{}75.69} & \textbf{\footnotesize{}56.36}\tabularnewline
\hline 
% \hline
% \cline{1-2} 
% \cmidrule(r){1-2}
\centering{}{\footnotesize{}R$\rightarrow$R} & \centering{}{\footnotesize{}Oracle} & {\footnotesize{}0.0680} & {\footnotesize{}72.78} & {\footnotesize{}89.66} & {\footnotesize{}41.94}\tabularnewline
% \hline 
\thickhline
\end{tabular}
\vspace{-1.2em}
\label{tab:flot}
\end{table}
%%%%%%%%%%%%%%%%%%%%%%%%%%%%%%%%%%%%%%%%%%%%%%%%%%%% FLOT %%%%%%%%%%%%%%%%%%%
% \linespread{1.0}
\renewcommand\arraystretch{1.0}

\vspace{-2mm}
\subsection{Ablation Study}
\label{subsec:ablation}
\vspace{-1mm}
We conduct ablation studies with HPLFlowNet as the baseline to evaluate the contribution of  individual components of our framework in the overall performance.

\vspace{0.5mm}
\noindent
\textbf{Effectiveness of DR and CR.} We investigate the effectiveness of the two key modules (\textit{i.e.,} DR and CR) in our framework by progressively adding them. In  Tab.~\ref{tab:DA}, Ours (E)
indicates our baseline using End-Point Consistency without DR and CR. 
Addition of DR and CR are shown with +D and +C in the table. It can be seen that DR consistently boosts the performances on all source-target dataset pairs. By further adding CR, our framework  achieves 39.4\% to 56.8\% improvement from FT3D to real datasets and 15.21\% to 45.28\% improvement from GTA-SF to real datasets.  This demonstrates explicit  contributions of DR and CR in our framework. 
Since there is a larger domain gap  between FT3D and real data, DR and CR are able to bring more improvements for the transferred models.

\vspace{0.5mm}
\noindent
\textbf{Effectiveness of Mean Teacher.} Our framework adopts mean teacher to provide pseudo labels for the unlabeled target domain. To verify the effectiveness of our teacher model, we conduct experiments by  replacing it with a model identical to the student by setting $\alpha$ in EMA to 0. Tab.~\ref{tab:MT} shows that the performance of our framework drops  after removing the teacher model. This identifies the contribution of mean teacher to provide positive supervision.

\vspace{0.5mm}
\noindent
\textbf{Effectiveness of Asymmetric Transformation.} We propose Asymmetric Transformation ($\mathcal{AT}$) as the augmentation strategy for the student model in our framework. $\mathcal{AT}$ transforms the first frame of input point cloud pairs with stochastic augmentations. Tab.~\ref{tab:AT} conducts an ablation study to evaluate the efficacy of $\mathcal{AT}$.
As compared to Symmetric Transformation ($\mathcal{ST}$) using the same augmentation for both input frames, $\mathcal{AT}$ enables better performance.
We analyze the effects of different augmentations in $\mathcal{AT}$. Compared with translation ($\mathcal{T}$) or translation+rotation ($\mathcal{T}+\mathcal{R}$), using rotation ($\mathcal{R}$) only yields the best results. This makes rotation a more suitable augmentation for SRSFE since it brings realistic scene flow in accord with motor steering.

% \linespread{0.8}
\begin{table}[t]
\caption{Contribution of Mean Teacher (MT) in our framework. $\downarrow$ and $\uparrow$ respectively indicate negative and positive polarity.}
\vspace{-2mm}
\centering{}%
\begin{tabular}{>{\centering}m{20mm}|>{\centering}m{10mm}>{\centering}m{10mm}>{\centering}m{10mm}>{\centering}m{10mm}}
% \hline 
\thickhline
\multirow{1}{18mm}{\centering{}{\footnotesize{}Settings}} & {\footnotesize{}EPE$\downarrow$} & {\footnotesize{}AS$\uparrow$} & {\footnotesize{}AR$\uparrow$} & {\footnotesize{}Out$\downarrow$}\tabularnewline
\hline 
\multirow{1}{18mm}{\centering{}{\footnotesize{}without MT}} & {\footnotesize{}0.0768} & {\footnotesize{}51.17} & {\footnotesize{}85.72} & {\footnotesize{}53.81}\tabularnewline
\multirow{1}{18mm}{\centering{}{\footnotesize{}with MT}} & \textbf{\footnotesize{}0.0683} & \textbf{\footnotesize{}58.57} & \textbf{\footnotesize{}87.98} & \textbf{\footnotesize{}47.40}\tabularnewline
% \hline 
\thickhline
\end{tabular}
\vspace{-0.6em}
\label{tab:MT}
\end{table}
% \linespread{1.0}
% \linespread{0.8}
\begin{table}[t]
\caption{Ablation on Asymmetric Transformation ($\mathcal{AT}$) and augmentations. Symmetric Transformation ($\mathcal{ST}$) has same transformation for both frames. $\mathcal{T}$ \& $\mathcal{R}$ mean Translation \& Rotation. $\downarrow$ and $\uparrow$ respectively indicate negative and positive polarity.}
\vspace{-2mm}
\centering{}%
\begin{tabular}{>{\centering}m{8mm}>{\centering}m{8mm}|>{\centering}m{10mm}>{\centering}m{10mm}>{\centering}m{10mm}>{\centering}m{10mm}}
% \hline 
\thickhline
\multicolumn{2}{c|}{{\footnotesize{}Methods}} & {\footnotesize{}EPE$\downarrow$} & {\footnotesize{}AS$\uparrow$} & {\footnotesize{}AR$\uparrow$} & {\footnotesize{}Out$\downarrow$}\tabularnewline
\hline 
\multicolumn{2}{c|}{{\footnotesize{}$\mathcal{ST}$}} & {\footnotesize{}0.0988} & {\footnotesize{}37.72} & {\footnotesize{}73.05} & {\footnotesize{}59.63}\tabularnewline
\hline 
\multirow{3}{8mm}{\centering{}{\footnotesize{}$\mathcal{AT}$}} & \centering{}{\footnotesize{}$\mathcal{T}$} & {\footnotesize{}0.0719} & {\footnotesize{}54.84} & {\footnotesize{}88.31} & {\footnotesize{}47.77}\tabularnewline
\cline{2-6} 
 & \centering{}{\footnotesize{}$\mathcal{T}+\mathcal{R}$} & {\footnotesize{}0.0702} & {\footnotesize{}55.28} & {\footnotesize{}88.27} & {\footnotesize{}47.73}\tabularnewline
\cline{2-6}
 & \centering{}{\footnotesize{}$\mathcal{R}$} & \textbf{\footnotesize{}0.0683} & \textbf{\footnotesize{}58.57} & \textbf{\footnotesize{}87.98} & \textbf{\footnotesize{}47.40}\tabularnewline
% \hline 
\thickhline

\end{tabular}
\vspace{-2mm}
\label{tab:AT}
\end{table}

\begin{figure}[t]
\vspace{-1mm}
  \centering
  \includegraphics[width=1\linewidth]{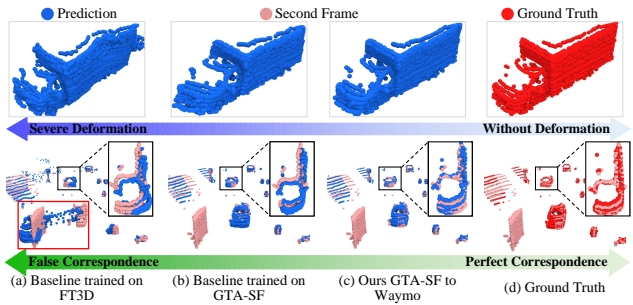}
  \vspace{-6mm}
  \caption{Qualitative comparisons on Waymo dataset. (a) Baseline trained on FT3D  seriously deforms objects and provides incorrect correspondence - red box. (b) Replacing FT3D with GTA-SF results in better predictions, but deformation and misalignment still exist - black boxes. (c) Our UDA method with GTA-SF achieves the best results by incorporating DR and CR.}
  \label{fig:def_cor}
  \vspace{-5mm}
\end{figure}

\vspace{0.5mm}
\noindent
\textbf{Analysis of Deformation and Correspondence.} Based on the observation that shape deformation and correspondence deviation are major problems for SRSFE, we make qualitative comparisons in Fig.~\ref{fig:def_cor} to show the  advantage of addressing them. It can be seen that the baseline trained on FT3D faces severe shape deformation and correspondence deviation when directly transferred to Waymo due to large domain gap. Replacing FT3D with GTA-SF gives a significant improvement on Waymo, though  the above two problems are only partially addressed. After applying our framework for unsupervised domain adaptation we achieve the  best results by explicitly addressing the mentioned problems.

%%%%%%%%%%%%%%%%%%%%%%%%%%%%%%

%%%%%%%%%%%%%%%%%%%%%%%%%%%%%%%%%%%%%

\section{Conclusion and Limitation} 
\label{sec:conclusion}
\vspace{-1mm}
We investigated synthetic-to-real scene flow estimation (SRSFE) on point clouds and addressed two major challenges for the task. First, based on the observation that a large domain gap exists between the existing synthetic datasets and real-world scenarios, we build a more realistic dataset for SRSFE using GTA-V engine. 
Second, to further reduce the domain gap for the induced computational models, we devise a mean-teacher-based framework for domain adaptation in scene flow estimation. Our framework incorporates end-point consistency during training with deformation regularization and correspondence refinement. We conclusively establish the reduced domain gap between our and real data by a quantitative evaluation on three real datasets. We also demonstrate a remarkable domain gap reduction over the existing baselines with our data and proposed framework. For pragmatic reasons, we employ rigidity assumption in our dataset. This only provides an approximation to non-rigid objects (\textit{e.g,} pedestrians). Nevertheless, this does not result in observable performance degradation, yet it provides computational advantages.  

\textit{We do not forsee any ethical concerns with the dataset and approach proposed in this manuscript, such as sensitive personal information, and bias against gender or race.}

% \noindent
% \textbf{Acknowledgement:} This work was partially supported by the National Natural Science Foundation of China (No. 61972435, U20A20185).

\balance

%%%%%%%%% REFERENCES
{\small
\bibliographystyle{ieee_fullname}
\bibliography{main}
}

\end{document}

% \title{Supplementary Material}

% \setcounter{section}{0}
% \setcounter{table}{0}
% \setcounter{figure}{0}
\def\thesection{Appendix \Alph{section}}

\thispagestyle{empty}

\vspace{-1mm}
\section{Implementation Details\label{sec:implementation-details}}
\subsection{Scene Flow Annotation\label{sec:scene flow generation}}
Scene flow annotation strategies are different for three real-world datasets (\textit{i.e.,} Waymo \cite{jund2021scalable, sun2020scalability}, Lyft \cite{Lyft} and KITTI \cite{menze2015object, menze2015joint}). The details are described as follows.

%Scene Flow Annotation / Generation (for waymo, how to generate scene flow... for Lyft... for KITTI...)
\vspace{2mm}
\noindent
\textbf{Waymo } The scene flow annotation of Waymo dataset \cite{jund2021scalable, sun2020scalability} is bootstrapped from tracked bounding boxes of objects. 
% expanding Waymo Open Dataset (WOD) \cite{sun2020scalability} with scene flow labels, which are bootstrapped from tracked bounding boxes of objects.
% It includes 158,081 training frames and 39,987 validation frames with scene flow annotation. 
% For efficiency, we choose half of validation frames for evaluation.
In fact, it provides scene flow annotations \textit{after compensating for ego-motion} (\textit{i.e.,} the movement of LiDAR).
However, this leads to the discarding of relative motion caused by LiDAR moving, so that the scene flow cannot translate the first frame to align with the second frame, which is different from previous datasets \cite{menze2015object, menze2015joint, mayer2016large}.

To enable existing models to be applied on Waymo, we follow the same setting as previous datasets \cite{menze2015object, menze2015joint, mayer2016large} and \textit{retrieve ego-motion}.
For a pair of point cloud frames with original scene flow vectors $(\mathbf{P}^s, \mathbf{P}^t, \mathbf{F}^{0})$, we first load each frame's transformation matrix $(\mathbf{T}^s, \mathbf{T}^t)$, which represents relative rotation and translation from world coordinates to LiDAR coordinates.
Subsequently, we retrieve ego-motion and compute scene flow vectors as
\begin{equation}
\mathbf{F}=\mathbf{P}^{s}-(\mathbf{P}^{s}-\Delta_{t}\mathbf{F}^{0})\cdot(\mathbf{T}^{s})^{-1}\cdot\mathbf{T}^{t},
\end{equation}
where $\Delta_{t}$ is the time interval between two frames. Since the scene flow annotation provided by Waymo is in form of speed, we first multiply it by $\Delta_{t}$ to get motion vectors.
% Note that Waymo takes the second frame as the source frame to compute scene flow, which can be used as an estimation of per-point future movement of the second frame.
Before training, we adopt a similar way as KITTI to remove ground points by height ($<0.3m$) and extract points in the front view.
Then we remove points by distance ($>60m$).
Moreover, we transform point clouds into the same coordinate system as KITTI by global rotation.

\vspace{2mm}
\noindent
\textbf{Lyft } We adopt the same way as Waymo to generate scene flow annotations for Lyft \cite{Lyft}.
Specifically, for each LiDAR frame in Lyft, it provides bounding boxes for foreground objects and two relative pose (\textit{i.e.,} rotation and translation) of sensor and car respectively.
In order to generate scene flow labels, we first leverage bounding boxes to make the alignment of objects with the same instance token from two consecutive frames.
After that, for points not included in bounding boxes, we compute ego-motion as their scene flow.
For training and evaluation, we adopt the same way as Waymo to remove undesired points.

\vspace{2mm}
\noindent
\textbf{KITTI } The point clouds and scene flow labels of KITTI Scene Flow dataset \cite{menze2015object, menze2015joint} are generated from stereo images using the same per-processing steps as \cite{gu2019hplflownet}.
As discussed in previous works \cite{gojcic2021weakly, jund2021scalable, baur2021slim}, KITTI is a semi-realistic dataset as \textit{it involves per-processing that inevitably changes real-world characteristics}. 
Our experimental results also show that synthetic-to-real performance degradation is less on KITTI.
Nevertheless, we also use KITTI as one of the target domain datasets to demonstrate the generalization ability of our method.

\subsection{Network Implementations}  
\noindent
$\bigcdot$ \textit{Input\quad} We subsample each frame into 8,192 points for all datasets as the input of the network.

\vspace{1mm}
\noindent
$\bigcdot$ \textit{Baselines\quad} For all three baselines (\textit{i.e.,} HPLFlowNet \cite{gu2019hplflownet}, FLOT \cite{puy2020flot} and PV-RAFT \cite{wei2021pv}), we use the default parameters as they described in their papers.

\vspace{1mm}
\noindent
$\bigcdot$ \textit{EMA\quad} $\alpha$ in EMA is set to 0.999 to update teacher model.

\vspace{1mm}
\noindent
$\bigcdot$ \textit{Asymmetric Transformation\quad} We adopt random global rotation along y axis as the augmentation strategy, which is the axis perpendicular to the ground.

\vspace{1mm}
\noindent
$\bigcdot$ \textit{Deformation Regularization\quad} We adopt the DBSCAN \cite{ester1996density} for clustering, which is a simple yet effective way to segment rigid objects since ground points have been removed before input into the network.

\vspace{1mm}
\noindent
$\bigcdot$ \textit{Correspondence Refinement\quad} We set $K$ to 6 to compute the Laplacian coordinates $\mathbf{L}^{1}$ and $\mathbf{L}^{2}$.

\vspace{1mm}
\noindent
$\bigcdot$ \textit{MMD\quad} We apply it to the outputs of DownBCL7 in HPLFlowNet for cross-domain feature alignment.

\vspace{1mm}
\noindent
$\bigcdot$ \textit{Training Strategy\quad} NVIDIA A10 with 24GB GPU memory is used for all of our experiments.
Since the teacher model is a temporal-ensembling of the student model, training from scratch with random initialized student model will lead to a meaningless teacher.
Therefore we pretrain HPLFlowNet on GTA-SF for 45 epochs, then we apply our method on it and train for 15 epochs to adapt to real-world datasets.

\begin{figure*}[t]
  \centering
%   \fbox{\rule{0pt}{1.8in} \rule{0.9\linewidth}{0pt}}
  \includegraphics[width=1\linewidth]{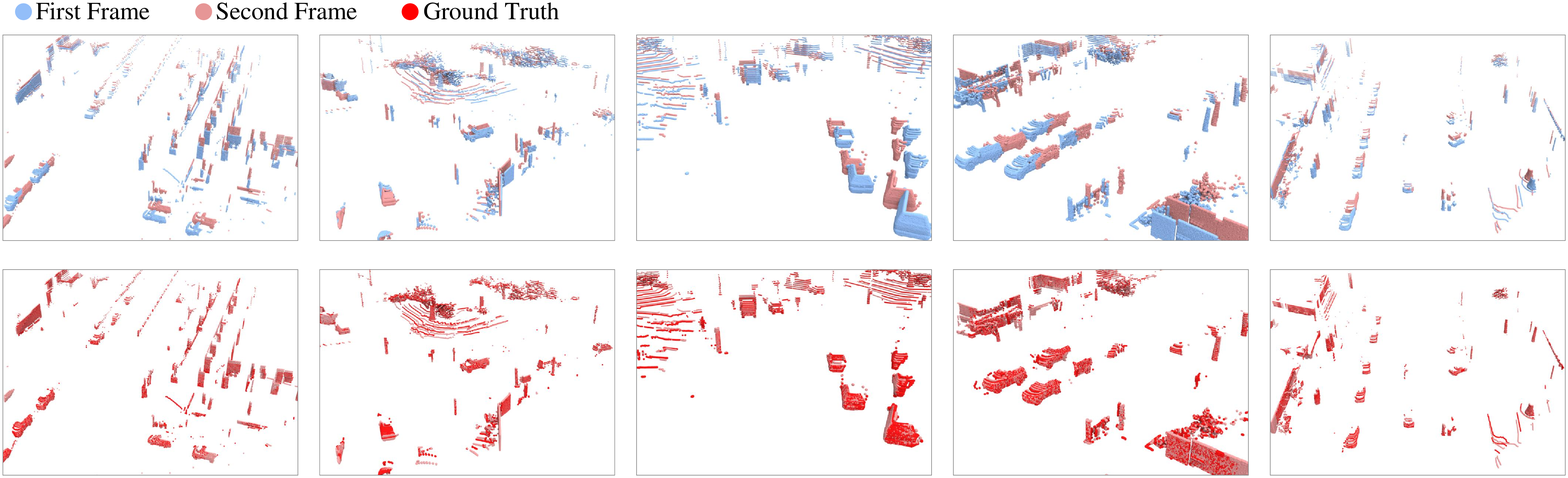}
  \vspace{-7mm}
  \caption{Illustration of point clouds and annotated scene flow in our GTA-SF. The first row shows two consecutive point cloud frames in each box, and the second row shows their corresponding ground truth obtained by adding generated scene flow to the first frame, which is well aligned with the second frame.}
  \vspace{-3mm}
  \label{fig:sup_GTA-SF}
\end{figure*}

\section{Detailed Analysis of GTA-SF}  
\label{suppl:Further}

We provide additional illustration of our proposed GTA-SF in Fig. \ref{fig:sup_GTA-SF} and further analyse the merits of it in the following.

\noindent
$\bigcdot$ \textit{Annotation\quad} We annotate scene flow for GTA-SF in a more realistic way.
In FT3D, two consecutive frames of point clouds are in point-wise correspondence, and the scene flow vectors $\mathbf{F}$ are directly computed as $\mathbf{F}=\mathbf{P}^t-\mathbf{P}^s$, where $\mathbf{P}^t$ and $\mathbf{P}^s$ are target and source point cloud.
However, such point-wise correspondence between two frames does not exist in real-world point clouds due to sensor movement and occlusion.
In our GTA-SF, we annotate scene flow in a similar way to Waymo. The main difference is that the bounding box in Waymo contains only one direction information, while we leverage all three directions of each entity to calculate more accurate scene flow in GTA-V.

\noindent
% \vspace{-2mm}
$\bigcdot$ \textit{Scene Diversity\quad} We build more realistic and diverse scenes to collect point clouds.
As mentioned in the main paper, our GTA-SF covers downtown, highway, streets and other driving areas.
Fig. \ref{fig:sup_GTA-SF} illustrates some typical point cloud frames with annotated scene flow in our GTA-SF. It can be seen that our GTA-SF covers a variety of vehicle driving scenarios. Moreover, our generated scene flow is able to accurately transform the first frame to match the second frame.

% explain why not in main text
\section{Additional Experimental Results}
\label{suppl:additional_experiments}
\subsection{Ablation on Parameters $\alpha$ and $K$}
\label{suppl:parameters}
\begin{figure}[ht]
\vspace{-2mm}
  \centering
%   \fbox{\rule{0pt}{1.8in} \rule{0.9\linewidth}{0pt}}
  \includegraphics[width=1\linewidth]{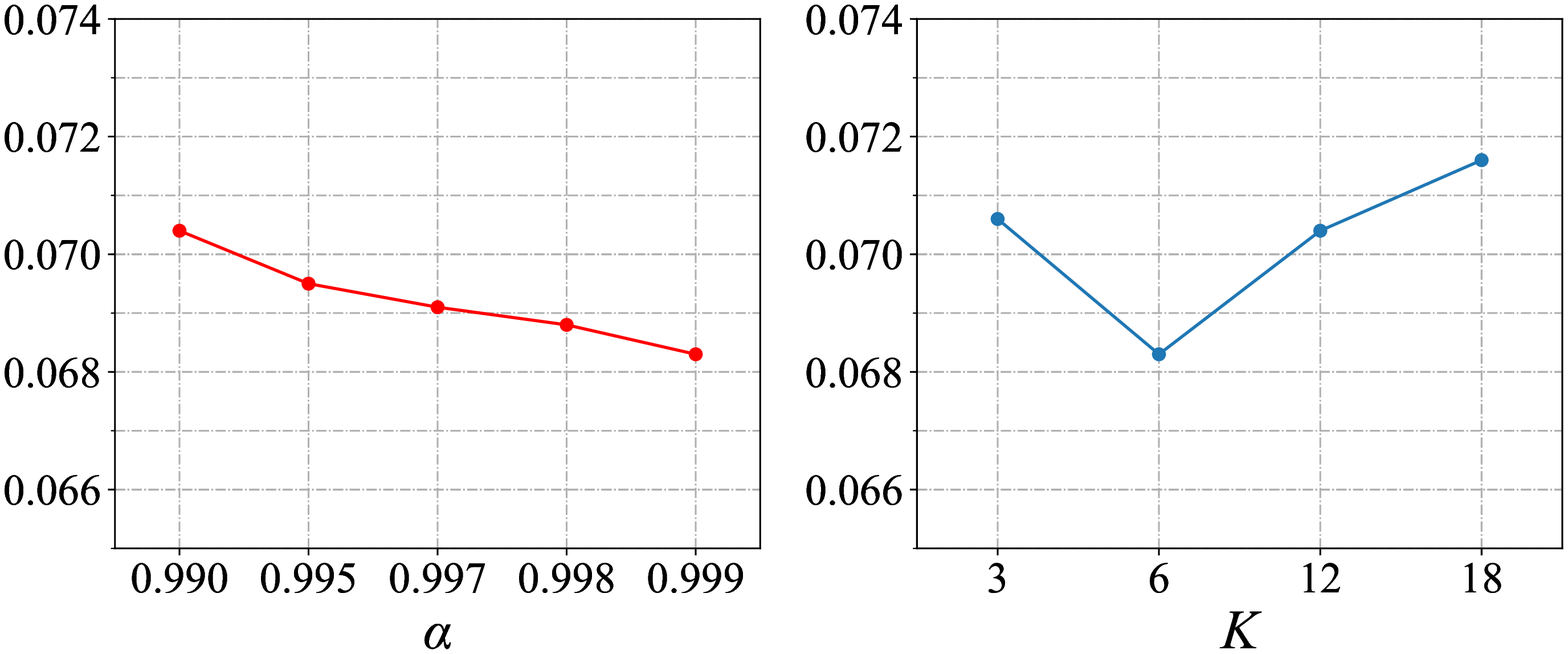}
  \vspace{-5mm}
  \caption{Effects of parameters $\alpha$ and $K$.}
  \vspace{-2mm}
  \label{fig:sup_parameters}
\end{figure}

We further conduct experiments on GTA-SF$\rightarrow$Waymo to analyse the effects of parameters $\alpha$ in EMA and $K$ in Correspondence Refinement (CR). $\alpha$ is a smoothing coefficient parameter controlling the update rate of the teacher model, and $K$ is the number of nearest neighbors used to calculate Laplacian coordinates.
Fig. \ref{fig:sup_parameters} draws the performance curves with different $\alpha$ and $K$.
We set $\alpha$ from 0.990 to 0.999 and $K$ from 3 to 18.
It shows that the best result is obtained when setting $\alpha$ to 0.999 and $K$ to 6.
In addition, our method is robust to parameters as there is less than 5\% relative performance fluctuation.

\subsection{Effectiveness of Ground Points Removal in GTA-SF}
\label{suppl:GPR}
As ground points are meaningless for scene flow and usually removed manually, in our GTA-SF, we propose to leverage gaming information for better Ground Points Removal (GPR). To validate our GPR is superior than general strategy, \textit{i.e.,} removing by height, Tab. \ref{tab:GPR} quantitative compares baseline results on Waymo using different GPR strategies for GTA-SF. It shows directly training on GTA-SF without GPR leads to poor performance. Compared to the common GPR by Height, our method removes ground points more effectively, facilitating training  scene flow estimators.

\label{sec:ab_gta}
\begin{table}[t]
\caption{Ablation on Ground Points Removal (GPR). $\downarrow$ and $\uparrow$ respectively indicate negative and positive polarity.}
% \vspace{-2mm}
\centering{}%
\begin{tabular}{>{\centering}m{20mm}|>{\centering}m{10mm}>{\centering}m{10mm}>{\centering}m{10mm}>{\centering}m{10mm}}
% \hline 
\thickhline
\multirow{1}{18mm}{\centering{}{\footnotesize{}Methods}} & {\footnotesize{}EPE$\downarrow$} & {\footnotesize{}AS$\uparrow$} & {\footnotesize{}AR$\uparrow$} & {\footnotesize{}Out$\downarrow$}\tabularnewline
\hline 
\multirow{1}{18mm}{\centering{}{\footnotesize{}without GPR}} & {\footnotesize{}0.1743} & {\footnotesize{}24.93} & {\footnotesize{}47.42} & {\footnotesize{}82.37}\tabularnewline
\centering{}{\footnotesize{}GPR by Height} & {\footnotesize{}0.1556} & {\footnotesize{}25.61} & {\footnotesize{}51.45} & {\footnotesize{}80.27}\tabularnewline
\centering{}{\footnotesize{}Our GPR} & \textbf{\footnotesize{}0.1061} & \textbf{\footnotesize{}32.35} & \textbf{\footnotesize{}66.21} & \textbf{\footnotesize{}65.35}\tabularnewline
% \hline 
\thickhline
\end{tabular}
\label{tab:GPR}
% \vspace{-2mm}
\end{table}

\subsection{Comparison with Self-Supervised Methods}
\label{suppl:self}
Recently, a number of self-supervised methods \cite{wu2020pointpwc, mittal2020just, li2021self, kittenplon2021flowstep3D} for scene flow estimation have emerged. They generally do not require any real-world ground-truth data, which is similar to our purpose. We reproduced the SOTA self-supervised method FlowStep3D \cite{kittenplon2021flowstep3D} on Waymo.
As shown in Tab. \ref{tab:SS}, (w/o self) means the model is pretrained on FT3D then directly tested on Waymo.
Our method shows superior performance.
Since self-supervised methods are finetuned on real data, 
our method explicitly accounts for key issues caused by domain gap, thus obtains larger improvements.

\begin{figure*}[!ht]
  \centering
%   \fbox{\rule{0pt}{1.8in} \rule{0.9\linewidth}{0pt}}
  \includegraphics[width=1\linewidth]{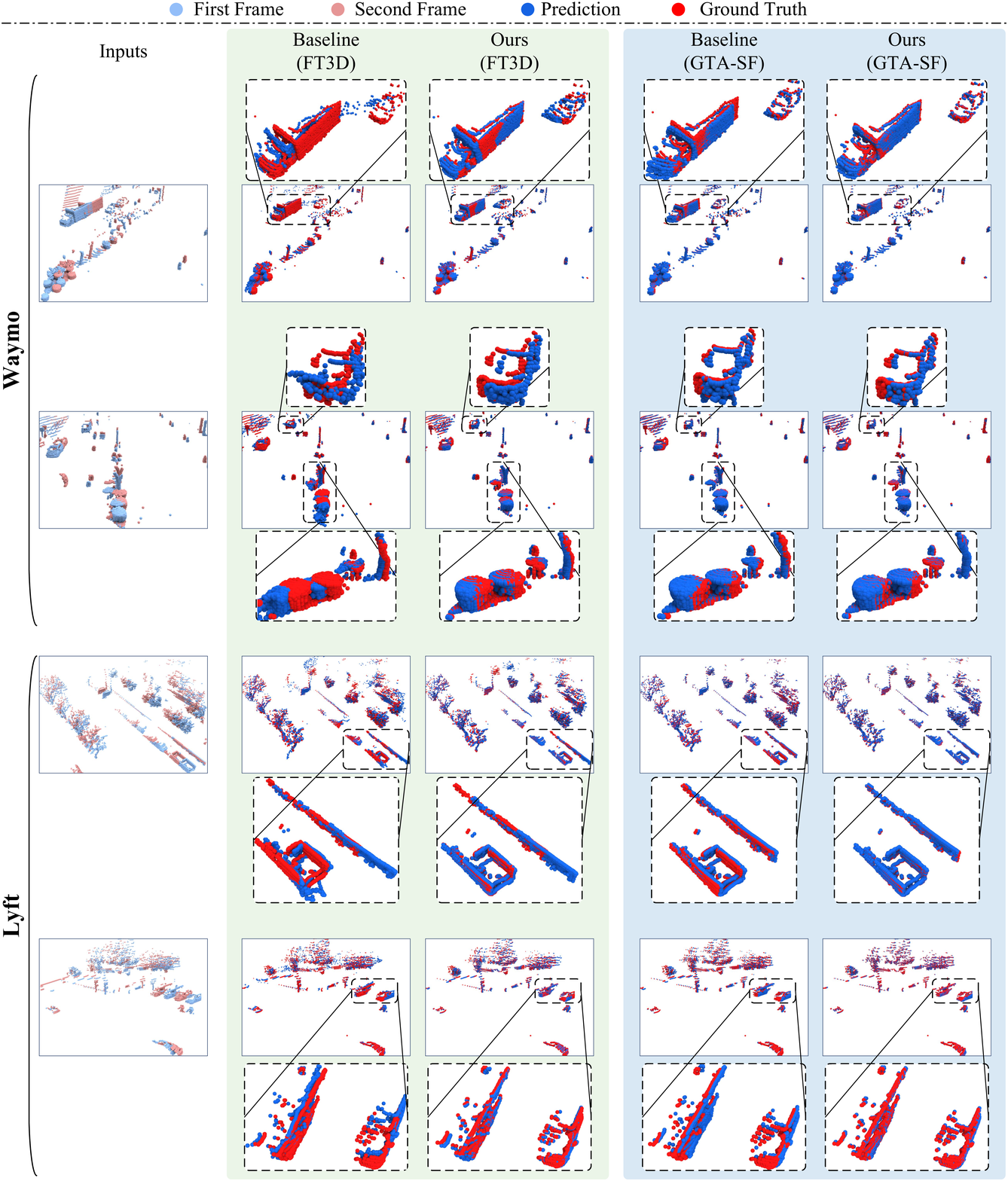}
  \vspace{-4mm}
  \caption{Qualitative comparison transferring from FT3D \cite{mayer2016large} and GTA-SF to Waymo \cite{sun2020scalability, jund2021scalable} and Lyft \cite{Lyft}. We compare the Baseline results trained on different synthetic datasets and the results after applying our proposed UDA method (Ours). Parentheses mark the used source domain dataset. The estimated scene flow is added to the first frame to get the prediction result for visualization, and performance can be judged by how well the prediction (\textcolor{blue}{blue}) align with the ground truth (\textcolor{red}{red}).}
  \label{fig:sup_qualitative_compare}
\end{figure*}

% \vspace{-3mm}
\begin{table}[t]
\caption{{Comparison with FlowStep3D on Waymo.}}
% \vspace{-3mm}
\centering{}%
\begin{tabular}{>{\centering}m{30mm}>{\centering}m{7.5mm}>{\centering}m{7.5mm}>{\centering}m{7.5mm}>{\centering}m{7.5mm}}
\hline 
\multicolumn{1}{>{\centering}m{30mm}}{\centering{}{\footnotesize{}Methods}} & {\footnotesize{}EPE$\downarrow$} & {\footnotesize{}AS$\uparrow$} & {\footnotesize{}AR$\uparrow$} & {\footnotesize{}Out$\downarrow$}\tabularnewline
\hline 
\multirow{1}{30mm}{\centering{}{\footnotesize{}FlowStep3D (w/o self)}} & {\footnotesize{}0.2476} & {\footnotesize{}35.31} & {\footnotesize{}61.77} & {\footnotesize{}68.69}\tabularnewline
\multirow{1}{30mm}{\centering{}{\footnotesize{}FlowStep3D (w/ self)}} & {\footnotesize{}0.1965} & {\footnotesize{}48.40} & {\footnotesize{}71.04} & {\footnotesize{}60.35}\tabularnewline
\hline 
\centering{}{\footnotesize{}Ours (w/o UDA)} & {\footnotesize{}0.2477} & {\footnotesize{}31.59} & {\footnotesize{}57.22} & {\footnotesize{}77.08}\tabularnewline
\centering{}{\footnotesize{}Ours (w/ UDA)} & \textbf{\footnotesize{}0.1251} & \textbf{\footnotesize{}48.87} & \textbf{\footnotesize{}78.40} & \textbf{\footnotesize{}57.29}\tabularnewline
\hline 
\end{tabular}
\label{tab:SS}
\end{table}

\subsection{Comparison with More Domain Adaptation Principles}
\label{suppl:DA_prin}
In addition to MMD and Self-Ensemble, we conduct experiments using adversarial-based DA approaches including DANN \cite{ganin2016domain} and ADDA \cite{tzeng2017adversarial}. We got 0.2515 EPE on FT3D$\rightarrow$Waymo. DANN showed a similar trend. We believe the primary goal of adverserial DA is to align latent features, while in scene flow estimation, explicit motion relationship between consecutive frames are important. We therefore adapted mean-teacher, since it enables using scene-flow outputs of momentum-updated teacher to guide learning of motion relationships.

\subsection{Analysis of the Effect of Synthetic Data on Real Data}
\label{suppl:S+R}
To further understand the value of synthetic data and know whether it is a good supplement to real ground-truth, we conduct the S+R$\rightarrow$R experiments and use additional GTA-SF data during training on Waymo.
Results (Tab.~\ref{tab:S+R}) show that additional GTA-SF data consistently improves oracle results, which indicates our GTA-SF is realistic and a good supplement for the real data.

\begin{table}[!bh]
% \vspace{-2mm}
\caption{{Comparison between R$\rightarrow$R and S+R$\rightarrow$R.}}
% \vspace{-3mm}
\centering{}%
\begin{tabular}{>{\centering}m{30mm}>{\centering}m{7.5mm}>{\centering}m{7.5mm}>{\centering}m{7.5mm}>{\centering}m{7.5mm}}
\hline 
\multicolumn{1}{>{\centering}m{30mm}}{\centering{}{\footnotesize{}Methods}} & {\footnotesize{}EPE$\downarrow$} & {\footnotesize{}AS$\uparrow$} & {\footnotesize{}AR$\uparrow$} & {\footnotesize{}Out$\downarrow$}\tabularnewline
\hline 
\multirow{1}{30mm}{\centering{}{\scriptsize{}Waymo$\rightarrow$Waymo}} & {\footnotesize{}0.0501} & {\footnotesize{}74.82} & {\footnotesize{}92.20} & {\footnotesize{}40.88}\tabularnewline
\multirow{1}{30mm}{\centering{}{\scriptsize{}GTA-SF+Waymo$\rightarrow$Waymo}} & \textbf{\footnotesize{}0.0482} & \textbf{\footnotesize{}76.39} & \textbf{\footnotesize{}92.53} & \textbf{\footnotesize{}39.06}\tabularnewline

\hline 
\end{tabular}
\label{tab:S+R}
\end{table}

\subsection{Qualitative Comparison}
\label{suppl:qualitative_compare}
In the main paper, we quantitative compare the domain adaptation performance on six source-target dataset pairs.
Due to the length limitation of the main paper, we further provide illustrations transferring from FT3D \cite{mayer2016large} and GTA-SF to Waymo \cite{sun2020scalability, jund2021scalable} and Lyft \cite{Lyft} to demonstrate the superiority of our method and GTA-SF dataset. 
As shown in Fig. \ref{fig:sup_qualitative_compare}, we qualitatively compare the performance by adding the estimated scene flow to the first frame, which is called prediction (colored in blue).
The prediction and ground truth are drawn in the same frame, so that the accuracy of scene flow prediction can be judged by how well the prediction align with the ground truth.
By zooming in on local details, we show that the prediction of our method is more consistent with the ground truth than baseline.
Moreover, the baseline results using GTA-SF as the source domain are better than those using FT3D, which verifies the significance of our proposed GTA-SF.
Note that we omit the visual comparisons on KITTI \cite{menze2015object, menze2015joint} since the qualitative improvement on KITTI is relative slight.
Nevertheless, our quantitative comparisons in the main paper demonstrate our method achieve consistent performance on KITTI.

% \newpage

\balance
%%%%%%%%% REFERENCES
{\small
\bibliographystyle{ieee_fullname}
\bibliography{supplementary}
}